\documentclass[final,5p,times,twocolumn]{elsarticle}

 \usepackage{CJKutf8} 
 \usepackage{lscape}
 \usepackage{setspace}
\usepackage{amssymb}
\usepackage{amsmath}
\usepackage{booktabs}
\usepackage{color}
\usepackage{float}

\graphicspath{{graph/}}
\usepackage{hyperref}

\usepackage{lipsum}
\usepackage{graphicx}
\usepackage[dvipsnames]{xcolor}
\usepackage{CJKutf8}
\usepackage{xcolor}
\usepackage{subcaption}
\usepackage{float}
\usepackage{booktabs}
\usepackage{array}
\usepackage{threeparttable}
\usepackage{multirow}

\usepackage{comment}
\usepackage{float}

\begin{document}
\begin{CJK*}{UTF8}{gbsn} 
\begin{frontmatter}

\title{Automated evaluation of LLMs for effective machine translation of Mandarin Chinese to English} 

\author{Yue Zhang\fnref{fn1}} 

\author{Rodney Beard\fnref{fn2}} 

\author{John Hawkins\fnref{fn2}} 
\author{Rohitash Chandra\fnref{fn1}} 


\affiliation[fn1]{organization={Transitional Artificial Intelligence Research Group, School of Mathematics and Statistics, UNSW Sydney},
            city={Sydney},
            country={Australia}}
            
\affiliation[fn1]{organization={Centre for Artificial Intelligence and Innovation, Pingla Institute},
            city={Sydney},
            country={Australia}}
\begin{abstract}
Although Large Language Models (LLMs) have exceptional performance in machine translation, only a limited systematic assessment of translation quality has been done. The challenge lies in automated frameworks, as human-expert-based evaluations can take an enormous amount of time, given the fast-evolving LLMs and the requirement for a diverse set of texts used for fair assessments of translation quality. In this paper, we utilise an automated machine learning framework featuring semantic and sentiment analysis to assess Mandarin Chinese to English translation using Google Translate and LLMs, including GPT-4, GPT-4o, and DeepSeek. We compare original and translated texts in various classes of high-profile Chinese texts, which include novel texts that span modern and classical literature, as well as news articles. As the main measures of evaluation, we utilise novel similarity metrics for comparing the quality of translations with the use of LLMs and get it further evaluated by an expert human translator. Our results indicate that the LLMs perform well in news media translation, but show divergence in their performance when applied to literary texts. Although GPT-4o and DeepSeek demonstrated better semantic conservation in complex situations, DeepSeek demonstrated better performance in preserving cultural subtleties and grammatical rendering. Nevertheless, the challenges in managing translation, maintaining culture-specificities, classical references and figurative expressions remain an open problem for all the respective models. 

\end{abstract}




\end{frontmatter}
\section{Introduction}
\label{sec:intro}

Machine translation is a field within Natural Language Processing (NLP) \citep{vaswani2017attention} that has undergone an evolutionary journey from rule-based approaches to statistical \citep{koehn2009statistical} and deep learning models
\citep{bahdanau2014neural,wu2016google,sutskever2014sequence}. The emergence of Large Language Models (LLMs) \citep{mann2020language} \citep{radford2019language,devlin2019bert} marks a new era for machine translation. The advent of GPT-4 has brought machine translation performance to the level of junior human translators, achieving results comparable to Google Translate in high-resource language pairs, such as Chinese-English translation \citep{jiao2023chatgpt}. Although GPT-4's performance is comparable to human translators, it still lags behind intermediate and advanced translators, with considerable performance gaps, particularly in specialised domains \citep{yan2024benchmarking}.

Google Translate  \citep{turovsky2016ten} has dominated the global machine translation market with its statistical models and subsequent deep learning model upgrades \citep{johnson2017google}. However, multiple studies have identified its limitations in Chinese translation: Google Translate cannot maintain contextual connections in Chinese-English translation and cannot accurately translate Chinese idioms (ch\'{e}ngy\v{u} 成语). These findings highlight the performance differences among various machine translation systems when processing Chinese texts \citep{hassan2018achieving,proietti2025has,toral2018level}. Recent developments in LLMs \cite{annepaka2025large} have transformed the landscape of machine translation, particularly for language pairs involving significant linguistic and cultural differences, such as Mandarin Chinese and English \citep{hendy2023good}. This has brought a new development in the area of human-quality translation through GPT-4, GPT-4o, and DeepSeek, with research showing that all three models are equally comparable to junior-level human translators \citep{yan2024gpt}. 

The task of machine translation between Chinese and English has limitations due to fundamental linguistic structures, writing systems, and cultural differences \citep{hassan2018achieving}. Mandarin Chinese is a nontypical logographic language with a low degree of morphological inflection in comparison to the high morphology of the English language \citep{yang2023performance}. Morphological inflection refers to the modification of words through affixes to express grammatical categories such as tense, number, person, and case; for instance, English employs suffixes like \textit{-s} for plural nouns and third-person singular verbs, \textit{-ed} for past tense, and \textit{-ing} for progressive aspect, whereas Chinese primarily relies on word order and functional particles to convey these grammatical relations \citep{li1989mandarin}. These variations are also compounded with the fact that Chinese writing is diverse, with ancient forms of vernacular Chinese surfacing in the Qing Dynasty to the modern form of journalistic writing that is typified by particular political and cultural expressions available today in communication domains \citep{yu2013cultural}. The intricacy of translations being done of cultures' inherent expressions, e.g., four-character idioms (ch\'{e}ngy\v{u}成语 ) and classical maxims （s\'{u}y\v{u}俗语 ） creates serious problems and challenges to modern machine translation systems that are unable to deliver sufficient support in these matters \citep{qiang2023chinese}.

Over the last several years, neural network-based systems have brought a revolution to tasks in NLP, and Transformer-based models perform optimally at a variety of language understanding tasks and benchmarks of language understanding \citep{ashish2017attention}.  BERT (Bidirectional Encoder Representations from Transformers)  is a prominent NLP model in contextual language representation \cite{devlin2019bert}, which is a pretrained model that creates dynamic, contextual representations. BERT consider both right and left context simultaneously to buy representation information and details of them at the same time as the text is processed in both directions \citep{rogers2021primer}. This two-way pre-training system allows BERT to better capture linguistic peculiarities, which is useful in applications where semantic understanding is required, such as machine translation evaluation and cross-lingual understanding \citep{zhang2020semantics}\citep{sun2022bort}.
  
The assessment of machine translation systems requires sophisticated evaluation with more advanced metrics, including semantic and sentiment analysis \citep{wang2024evaluation}, which can complement classic metrics such as Bilingual Evaluation Understudy (BLEU) \cite{papineni2002bleu}. Even though BLEU scores have been popular in the context of quantitative measurement, more recent studies have indicated that embedding-based metrics, such as  BERT-Score, are also better correlated with human rating, especially between morphologically different language pairs such as Chinese-English \citep{cui2024automated}. Sentiment analysis provides another layer of evaluation since it helps to identify the subtleties of emotion and feelings that are often lost in translation \citep{chandra2025evaluation}.

This study builds on the previous studies of Shukla et al. \citep{shukla2023evaluation} who report an automated translation evaluation analysis of Sanskrit through Google Translate sentiments and semantic analysis using the corpus of the Bhagavad Gita. Similarly, Chandra and Kulkarni \citep{chandra2025evaluation} also used sentiment analysis and semantic analysis to compare Sanskrit-to-English translations done by three human participants. Our study relies on the study by Cui and Liang \citep{cui2024automated}, which presented an automated scoring system based on BERT-based models to check the quality of Chinese English translation. Their study used multilingual BERT embeddings to find semantic similarities between source texts and translations and achieved remarkable similarity with human ratings for a variety of genres, including news articles, literary works, and technical documentation.

Although Neural Machine Translation (NMT) has made great progress \cite{lo2025neural}, current studies have shown that there are gaps in the assessment of LLMs for the translation of various types of Chinese texts. Earlier experiments have largely concentrated on individual domains or text types, but comparative evaluation among genres is understudied \citep{chen2024large}. Furthermore, although Google Translate has been extensively studied for various language pairs \cite{mirzaeian2023google}, comprehensive comparisons between Google Translate and state-of-the-art LLMs such as GPT-4 and DeepSeek for Chinese-English translation remain scarce. The cultural and historical knowledge required for translating classical Chinese literature, the narrative complexity of modern novels, and the specialised terminology of news texts each present unique challenges that warrant systematic investigation \citep{gao2024machine}.

This study addresses the research gaps by conducting a comprehensive evaluation of Google Translate, GPT-4, GPT-4o and DeepSeek for Mandarin Chinese to English translation across three representative text categories.  We utilise a machine learning framework that uses semantic and sentiment analysis to compare original and translated texts. We select ``A Dream of Red Mansions''(H\'{o}ngl\'{o}um\`{e}ng《红楼梦》) \cite{cao1994hongloumeng} to represent classical vernacular literature, Mo Yan's ``Red Sorghum''(H\'{o}ngg\={a}oli\'{a}ng《红高粱》) \cite{mo2008honggaoliang} for modern Chinese novels and articles from ``Global Times''(Hu\'{a}nqi\'{u} Sh\'{i}b\`{a}o《环球时报》) \cite{tan2014huanqiushibao} for contemporary news discourse. Such a choice helps to study the ways various machine translators address the specific linguistic, stylistic and cultural peculiarities of specific genres. We utilise a multidimensional machine-learning-based evaluation framework that simultaneously uses BERT-scoring and cosine similarity as semantic accuracy assessment metrics, as well as sentiment analysis as emotional fidelity assessment metrics. The comparison of machine translation outputs to expert translation by human translators helps us to clearly understand the linguistic performance of contemporary translation systems and their ability to retain cultural and emotional connotations.

The structure of this paper is as follows: Section 2 presents the research background, and Section 3 presents the research methodology with details about the evaluation framework. Section 4 presents the research results, Section 5 provides a discussion, and finally, Section 6 presents the conclusions.

\section{Background}
\label{sec:background}

\subsection{Challenges in Chinese-English Translation}
\label{subsec:challenges}

The translation of Mandarin Chinese to English is challenged by issues not present in the translation of European language pairs \citep{liu2024impact}. The key problem is word segmentation, i.e. division of a sequence of characters into individual words \citep{chang2008optimizing}. Contrary to the English language, where words are delimited by spaces, the Chinese text is composed of a continuous set of characters without definite boundaries between words \citep{hsu2000effects,bai2008reading}. The Chinese sentence {m\v{e}i gu\'{o} hu\`{i} b\`{u} t\'{o}ng y\`{i}}美国会不同意 can be segmented as either {m\v{e}i gu\'{o}}美国/{hu\`{i}}会/{b\`{u} t\'{o}ng y\`{i}}不同意 (The US will not agree) or {m\v{e}i}美/{gu\'{o} hu\`{i}}国会/{b\`{u} t\'{o}ng y\`{i}}不同意 (The US Congress does not agree), creating ambiguity that readers must resolve through context \citep{li2009segmentation}. Zhang et al. \citep{zhang2008chinese} reported that incorrect parsing of Chinese words could have 66.7\% of translation errors in major translation programs. These parsing issues in early-stage processing persist throughout the entire translation process \citep{huang2007chinese,zhang2016transition,zhao2006effective}. From a linguistic typology perspective, although both Chinese and English are SVO (Subject-Verb-Object) languages, they exhibit essential differences in their grammar \citep{li1989mandarin}. Chinese lacks tense conjugation forms and requires contextual inference, which machine translation systems struggle to process accurately \citep{cui2023comparative}. Chinese uses aspect markers such as ``了'' (le) and ``过'' (gu\`{o}) to express temporal concepts rather than verb tense changes as in English \citep{lin2003temporal}. This distinction between ``aspect-prominent'' and ``tense-prominent'' languages increases translation complexity \citep{collart2021processing,smith2009temporal,mcenery2010corpus}.

\subsection{Sentiment and Semantic Analysis}
\label{subsec:framework}

Combining sentiment and semantic analysis offers a multidimensional approach 
to assessing translation quality. Sentiment analysis, which computationally 
identifies opinions, sentiments, and emotions from text \citep{pang2008opinion,liu2022sentiment}, when combined with semantic analysis, provides a framework that evaluates not only lexical accuracy but also the preservation of emotional tone and semantic meaning across languages. Sentiment analysis can be conducted at the sentence level and the document level. Sentence-level sentiment analysis explains the stance of individual sentences, while document-level analysis examines the meaning of the entire text and the consistency of emotions within it \citep{socher2013recursive}. The BERT model \citep{devlin2019bert} offers a rich contextual analysis with the help of bidirectional encoding and multi-headed attention, allowing it to semantically represent difficult content like idioms and literary rhetoric \citep{kenton2019bert,goldberg2016primer,tenney2019bert}. In the case of semantic similarity evaluation, the MPNet model \citep{song2020mpnet} has achieved 90.4--90.8\% of translation matching to human translation \citep{reimers2019sentence}. The metric evaluation measures based on these neural models, such as BERTScore \citep{zhang2019bertscore} based on semantic similarity computation using BERT embeddings, BLEURT (Bilingual Evaluation Understudy with Representations from Transformers) \citep{sellam2020bleurt} based on learned semantic matching metrics using BERT as its basis, and MPNet based on cosine similarity scores for semantic matching assessments, demonstrate higher correlation with human evaluation than classical BLEU scores \citep{mathur2020tangled}.

\subsection{Evaluation Methods}
\label{subsec:innovations}

Conventional machine translation evaluation methods have been based on n-gram matching metrics such as BLEU \citep{papineni2002bleu}. However, studies indicate that neural network evaluation metrics improve human judgments by 0.1-0.2 points greater than conventional assessment metrics \citep{rei2020comet}\citep{ma2019results}\citep{freitag2021experts}. Although such sophisticated measurements can give a picture of the quality in general, they do not show fine-grained patterns of lexical representations and stylistic inclination of various translation systems. Therefore, this study complements the sentiment and semantic evaluations with trigram analysis \citep{brown1992class}, which can reveal unique characteristics in lexical selection and language patterns across different translation systems, particularly after removing stop words, allowing better focus on meaningful word combinations \citep{manning1999foundations}\citep{coleman2005introducing}\citep{chen1999empirical}.

The comprehensive evaluation framework used in this study is based on Chandra et al.  \citep{chandra2025evaluation}, which combines BERT-based sentiment analysis with MPNet-based semantic similarity assessment to provide multidimensional translation quality evaluation. This method is an improvement over single metric rating because it emphasises both preservation of emotion and semantic accuracy on varying text types at the same time \citep{chandra2025evaluation}. COMET (Crosslingual Optimised Metric for Evaluation of Translation) is also a neural-based evaluation that uses the XLM-RoBERTa encoder with a correlation of over 0.85 with human ratings \citep{wang2023findings}. This is an important multidimensional method of evaluation to know the strengths and weaknesses of various translation systems in the processing of various kinds of Chinese texts \citep{kocmi2021ship,lommel2014multidimensional,specia2018quality}. 

\section{Methodology}
\label{sec:methodology}

\subsection{Text Types}
\label{subsec:variability}

We select three representative types of Chinese texts to comprehensively evaluate the translation quality of large language models in different contexts. The selection of datasets was based on the diversity of linguistic features, cultural connotations, and historical backgrounds of the texts.

Depending on the type of Chinese text, the domain-specificity level in translation varies greatly \citep{saldanha2014research}. Compared with other content types, news articles have the lowest translation entropy and cognitive load \citep{carl2017sketch}. Literary translation needs more cultural background knowledge and creative imagination \citep{carl2017sketch}. The three types of texts used in our study are expected to present various translation challenges: 

\begin{itemize}
\item \textit{Modern news}: We selected \textit{Global Times}, which has fixed language structures and clear information source purposes, and is an appropriate translation environment for translation systems \citep{pan2023multimodality,xiao2015corpus,liu2018news}.  As one of China's major English-language media outlets, \textit{Global Times} features standardized linguistic characteristics and specific discourse systems in its news reporting \citep{wang2021chinese}. This study directly adopted the English versions published by the newspaper as expert translation references, ensuring the timeliness and accuracy of translations.

\item \textit{Classical literature }:  We selected \textit{A Dream of Red Mansions}, which has more than 3,000 included poems, complicated naming systems, and multi-layered narrative information, and thus presents translation challenges \citep{zhou2025translator}.  As the pinnacle of Chinese classical literature, \textit{A Dream of Red Mansions} retains the elegance of classical literary Chinese while integrating vernacular features from the Ming and Qing dynasties, containing numerous poems, idioms, and cultural metaphors \citep{zhouruchang2018}. This study adopted H. Bencraft Joly's English translation as the expert translation reference, which was first published in 1892-1893 and is one of the earliest complete English translations, renowned for its faithful preservation 

\item \textit{Modern fiction}: We selected \textit{Red Sorghum} as the source for modern fiction due to its popularity and translation availability.  Recent research indicated that Howard Goldblatt's English translation deleted 50,000 Chinese characters (13\% of the original amount), and 70\% of English readers failed to recognise the incompleteness of the translation \citep{wu2022sociological}. As the first Chinese writer to win the Nobel Prize in Literature, Mo Yan's works integrate magical realism with Chinese rural literary traditions, featuring vivid language with distinctive regional characteristics \citep{zhanghong1999}. Howard Goldblatt's translation is recognized as the authoritative English version of Mo Yan's works, and his long-term collaboration with Mo Yan ensures that the translation maintains both the literariness of the original work and adapts to English readers' reading habits \citep{goldblatt2014mutually}.
\end{itemize}




\subsection{Data Extraction and Processing}
\label{subsec:processing}

We processed the original Chinese texts with the following steps:
\begin{enumerate}
 
  \item  \textit{Text Extraction:} Complete texts of the selected chapters were extracted out of the digital material.

  \item  \textit{Format Standardisation :} Text encoding was standardised at UTF-8: this ensured that the traditional simplification issue with the conversion of the relevant texts was solved, but at the same time, the original paragraph structure was preserved \citep{chendeguang2021}.

  \item  \textit{Paragraph Segmentation:} The natural paragraph constituency of the original text was retained, which is important in ensuring that the contextual integrity was maintained. It has been found that paragraph processing leads to semantic preservation performance when compared to sentence processing in terms of semantic integrity preservation in research findings \citep{wong2012extending}.
 
\end{enumerate}

\subsection{LLM Evaluation Framework}
\label{subsec:Analytical Framework}

\begin{figure*}[htbp]
\centering
\includegraphics[width=\textwidth]{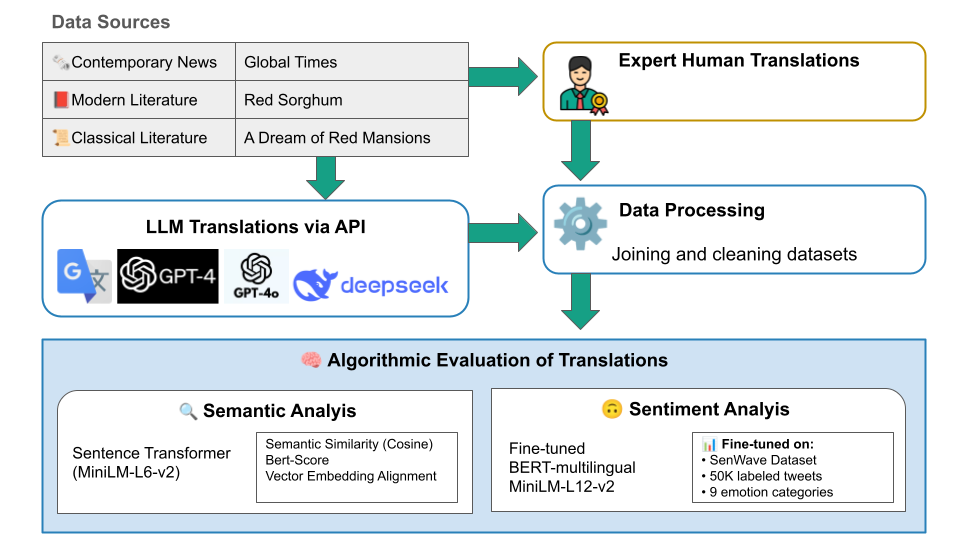}
\caption{Sentiment and Semantic analysis framework for comparison of translations by LLMs}


\label{fig:framework}
\end{figure*}

Our LLM evaluation framework for machine translation is based on works by Chandra et al. \citep{chandra2025evaluation} and Shukla et al. \citep{shukla2023evaluation} (Figure~\ref{fig:framework}), and amended to the specifics of Chinese-English translation. 

Our LLM-evaluation framework involves a series of stages, where Stage 1 entails retrieving sampled Chinese source and proficient translations. 

In Stage 2, we call various translation APIs to generate translations and organize the corresponding English translations in a structured format stored in a GitHub repository \footnote{\url{https://github.com/yuezhang0817/LLM-Translation-Madarin.git}}. We used three translation systems in generating English translations:

\begin{enumerate}
  
  \item \textit{Google Translate:} Invoked via Google Cloud Translation API (v3)  \footnote{\url{https://docs.cloud.google.com/translate/docs}}., without providing specific parameters to preserve the original state of translation \citep{ahmed2025evaluation}.

  \item \textit{GPT-4 and GPT-4o:} Invoked via OpenAI API  \footnote{\url{https://platform.openai.com/docs/overview}}. in models with specific parameters. In line with best practices, we picked which system to use: ``Please translate the following Chinese text into English, it should be in the original meaning and style to render the translation consistent \citep{jiao2023chatgpt}.

  \item  \textit{DeepSeek:} called through the DeepSeek API  \footnote{\url{https://api-docs.deepseek.com/}} and to ensure consistency in translation using LLM, we used the same system prompts as in GPT. \citep{liu2024deepseek}\citep{bi2024deepseek}.

\end{enumerate}

In Stage 3, we feature data preprocessing to  remove deactivated words to focus on meaningful vocabulary combinations, to reveal the unique characteristics of different translation systems in vocabulary selection and language expression.

In Stage 4, we perform n-gram analysis   (bigram and trigram) to compare the vocabulary selection and language patterns of human expert translations and machine translation systems (including Google Translate, DeepSeek, GPT-4 and GPT-4o) in processing different types of Chinese texts. Bigram analysis captures basic vocabulary pairing and naming habits, while trigram analysis reveals more complex narrative structures and context patterns across different text types.

In Stage 5, we use BERT-based models for paragraph-level sentiment classification, semantic similarity calculation, and statistical feature extraction. During model training, we fine-tune the model using the SenWave dataset \citep{yang2020senwave}, which contains 50,000 annotated multilingual tweets. The original dataset includes 10 sentiment categories (optimistic, thankful, empathetic, pessimistic, anxious, sad, annoyed, denial, official report, joking). Considering the characteristics of literary texts, we excluded the ``official report'' category related to COVID-19, ultimately using 9 sentiment categories for classification \citep{mohammad2010emotions}.

In Stage 6, we perform semantic analysis using an MPNet-based model to compute the cosine similarity between machine translations and human expert translations. This method evaluates semantic consistency across three text types. The cosine similarity scores range from 0 to 1, with 1 indicating perfect semantic alignment, enabling systematic comparison of translation fidelity across different systems and text genres.

In Stage 7, we perform result visualisation and evaluation, generate sentiment distribution charts, calculate comprehensive quality scores, and conduct statistical significance tests.

In Stage 8, we implement a comprehensive analysis by integrating semantic similarity scores with sentiment analysis results to examine the relationship between translation accuracy and emotional fidelity. This cross-dimensional evaluation reveals whether translations that are semantically "correct" also preserve the appropriate emotional tone.


As a model selection, the sentiment analysis in the present paper uses the sentence-transformers/ paraphrase-multilingual-MiniLM-L12-v2 model with strengths including multilingual support, cross-lingual performance, and semantic embedding capabilities \citep{reimers2020making}. 









\subsection{Metrics}

Our framework employs the BERT score as the main evaluation index in Stage X,  and the implementation is as follows:

\begin{enumerate}
\item Generate contextual embeddings for reference and candidate translations
\item Calculate token-level cosine similarity matrix
\item Use greedy matching algorithm to find optimal alignment
\item Calculate precision, recall, and F1 scores \citep{zhang2019bertscore}
\end{enumerate}

Studies show that BERT-Score achieves over 0.9 correlation with human judgment in Chinese-English translation evaluation \citep{cui2024automated}.

Furthermore, we use cosine similarity to measure paragraph-level semantic fidelity:

\begin{equation}
\text{similarity} = \frac{\mathbf{A} \cdot \mathbf{B}}{||\mathbf{A}|| \times ||\mathbf{B}||}
\end{equation}

where $\mathbf{A}$ and $\mathbf{B}$ are sentence embedding vectors of expert translations and machine translations, respectively \citep{arora2017simple}.

\section{Results}
\label{sec:results}

\subsection{N-Grams for Translation Analysis}
\label{subsec:data analysis}

\begin{table}[!t]
\centering
\caption{Top Bigrams from Global Times News Translations}
\label{tab:news_bigrams}
\small
\begin{tabular}{|l|c|l|c|}
\hline
\multicolumn{2}{|c|}{\textbf{Expert Translation}} & \multicolumn{2}{c|}{\textbf{GPT-4o}} \\
\hline
Bigram & Freq & Bigram & Freq \\
\hline
global times & 6 & xi jinping & 6 \\
china studies & 8 & economic trade & 7 \\
asian countries & 10 & asian countries & 9 \\
china central & 12 & china central & 12 \\
central asian & 21 & central asian & 19 \\
\hline
\multicolumn{2}{|c|}{\textbf{GPT-4}} & \multicolumn{2}{c|}{\textbf{Google Translate}} \\
\hline
Bigram & Freq & Bigram & Freq \\
\hline
middle east & 5 & economic trade & 9 \\
china central & 9 & china central & 10 \\
asian countries & 9 & asian countries & 13 \\
united states & 11 & central asian & 20 \\
central asian & 20 & united states & 28 \\
\hline
\multicolumn{2}{|c|}{\textbf{DeepSeek}} & \multicolumn{2}{c|}{} \\
\hline
Bigram & Freq & \multicolumn{2}{c|}{} \\
\hline
united states & 8 & \multicolumn{2}{c|}{} \\
asian countries & 9 & \multicolumn{2}{c|}{} \\
china studies & 10 & \multicolumn{2}{c|}{} \\
china central & 10 & \multicolumn{2}{c|}{} \\
central asian & 18 & \multicolumn{2}{c|}{} \\
\hline
\end{tabular}
\end{table}

We observe significant pattern differences in the translation of the Global Times, as shown in the bigrams (Table~\ref{tab:news_bigrams}).  The expert translation shows specific characteristics of geopolitical terms, such as `central asian' (21 times), `china central' (12 times) and `asian countries' (10 times), as well as the media source identifier `global times' (6 times), reflecting the professionalism and geographical concerns of the news discourse. GPT-4o is highly consistent with expert translations in core geopolitical terms (`central asian' 19 times, `china central' 12 times), but introduces the name of politicians and leaders `xi jinping' (6 times), showing its attention to news figures. Both GPT-4 and Google Translate frequently use `united states', indicating the emphasis on the narrative of China-United States relations, where the bigram frequency of Google Translate is obviously high, which may reflect its dependence on the framework of comparative international relations. While maintaining the core terms (`central asian' 18 times, `china central' 10 times), DeepSeek retains the academic expression `china studies' (10 times), reflecting its unique understanding of the original context.

\begin{table}[!t]
\centering
\caption{Top Bigrams from Dream of the Red Chamber Translations}
\label{tab:red_mansions_bigrams}
\small
\begin{tabular}{|l|c|l|c|}
\hline
\multicolumn{2}{|c|}{\textbf{Expert Translation}} & \multicolumn{2}{c|}{\textbf{GPT-4o}} \\
\hline
Bigram & Freq & Bigram & Freq \\
\hline
upon hearing & 9 & lady wang & 15 \\
lady chia & 11 & upon hearing & 15 \\
lost time & 12 & grandmother jia & 23 \\
madame wang & 13 & shi yin & 28 \\
dowager lady & 27 & yu cun & 40 \\
\hline
\multicolumn{2}{|c|}{\textbf{GPT-4}} & \multicolumn{2}{c|}{\textbf{Google Translate}} \\
\hline
Bigram & Freq & Bigram & Freq \\
\hline
lady wang & 12 & son jia & 8 \\
upon hearing & 12 & gave birth & 8 \\
grandma jia & 15 & feng su & 9 \\
yu cun & 21 & jia mu & 25 \\
shi yin & 36 & shi yin & 26 \\
\hline
\multicolumn{2}{|c|}{\textbf{DeepSeek}} & \multicolumn{2}{c|}{} \\
\hline
Bigram & Freq & \multicolumn{2}{c|}{} \\
\hline
said lady & 9 & \multicolumn{2}{c|}{} \\
lady xing & 10 & \multicolumn{2}{c|}{} \\
old lady & 12 & \multicolumn{2}{c|}{} \\
lady dowager & 13 & \multicolumn{2}{c|}{} \\
lady wang & 14 & \multicolumn{2}{c|}{} \\
\hline
\end{tabular}
\end{table}

The bigrams of \textit{A Dream of Red Mansions} (Table~\ref{tab:red_mansions_bigrams}) reveal that LLMs encounter certain difficulties with classical Chinese literature. The expert translations demonstrate sophisticated handling of character titles and relationships, with high-frequency bigrams such as `dowager lady' (27 times), `madame wang' (13 times), `lady chia' (11 times), and `upon hearing' (9 times), reflecting the hierarchical social structure and narrative rhythm of classical Chinese fiction. GPT-4o shows strong adaptation to classical naming conventions, with `yu cun' (40 times), `shi yin' (28 times), and `grandmother jia' (23 times) indicating its ability to recognize and maintain character name consistency, though the modernized term `grandmother jia' contrasts with the more classical `dowager lady' in expert translations. GPT-4 demonstrates similar character recognition patterns (`shi yin' 36 times, `yu cun' 21 times), but introduces variations such as `grandma jia' (15 times), suggesting a less formal register. Google Translate exhibits simplification tendencies with basic relationship terms such as `shi yin' (26 times), `jia mu' (25 times), and common action phrases (`gave birth' 8 times), reflecting a more literal approach to classical terminology. DeepSeek uniquely maintains classical formality through consistent use of aristocratic titles: `lady wang' (14 times), `lady dowager' (13 times), `old lady' (12 times), and `lady xing' (10 times), demonstrating particular sensitivity to the social hierarchy embedded in the original text.

\begin{table}[!t]
\centering
\caption{Top Bigrams from Red Sorghum Translations}
\label{tab:red_sorghum_bigrams}
\small
\begin{tabular}{|l|c|l|c|}
\hline
\multicolumn{2}{|c|}{\textbf{Expert Translation}} & \multicolumn{2}{c|}{\textbf{GPT-4o}} \\
\hline
Bigram & Freq & Bigram & Freq \\
\hline
detachment leader & 11 & sun wu & 12 \\
wang wenyi & 12 & grandpa luohan & 36 \\
black water & 17 & grandpa luo & 44 \\
uncle arhat & 55 & luo han & 48 \\
commander yu & 58 & commander yu & 61 \\
\hline
\multicolumn{2}{|c|}{\textbf{GPT-4}} & \multicolumn{2}{c|}{\textbf{Google Translate}} \\
\hline
Bigram & Freq & Bigram & Freq \\
\hline
wang wenyi & 16 & captain leng & 12 \\
luo han & 21 & sun wu & 14 \\
uncle luo & 26 & wang wenyi & 18 \\
uncle luohan & 44 & commander yu & 61 \\
commander yu & 63 & uncle luohan & 62 \\
\hline
\multicolumn{2}{|c|}{\textbf{DeepSeek}} & \multicolumn{2}{c|}{} \\
\hline
Bigram & Freq & \multicolumn{2}{c|}{} \\
\hline
japanese soldier & 12 & \multicolumn{2}{c|}{} \\
japanese soldiers & 13 & \multicolumn{2}{c|}{} \\
wang wenyi & 15 & \multicolumn{2}{c|}{} \\
commander yu & 61 & \multicolumn{2}{c|}{} \\
uncle arhat & 78 & \multicolumn{2}{c|}{} \\
\hline
\end{tabular}
\end{table}

Various translation models demonstrate understanding of Mo Yan's unique narrative style to different extents in the translation of his work, \textit{Red Sorghum} (Table~\ref{tab:red_sorghum_bigrams}.  The expert translation reveals distinctive character naming patterns, with high-frequency bigrams such as `commander yu' (58 times), `uncle arhat' (55 times), `black water' (17 times), and `wang wenyi' (12 times), reflecting the novel's focus on military figures and rural characters. The term `detachment leader' (11 times) emphasises the military organisational structure central to the narrative. GPT-4o shows adaptive character recognition with `commander yu' (61 times), maintaining consistency with expert translation, but introduces variations in character titles: `luo han' (48 times), `grandpa luo' (44 times), and `grandpa luohan' (36 times), indicating attempts to clarify familial relationships through multiple naming strategies. The presence of `sun wu' (12 times) suggests attention to historical military references. GPT-4 demonstrates the highest frequency for `commander yu' (63 times), showing strong alignment with the protagonist's role, yet exhibits significant fragmentation in translating secondary character names: the same character appears as `uncle luohan' (44 times), `uncle luo' (26 times), and `luo han' (21 times), revealing lack of systematic approach to kinship-title combinations and suggesting inconsistent decision-making across translation contexts. Google Translate shows concentrated attention to main characters (`uncle luohan' 62 times, `commander yu' 61 times), but the appearance of `captain leng' (12 times)—absent in other translations—likely represents misidentification of a character's military rank or conflation of different characters, exposing limitations in tracking narrative continuity and understanding the historical military hierarchy of 1930s rural China. DeepSeek uniquely emphasizes the war context through explicit military terms: `uncle arhat' (78 times, highest frequency), `commander yu' (61 times), `japanese soldiers' (13 times), and `japanese soldier' (12 times), demonstrating particular sensitivity to the novel's historical war narrative and antagonist forces.

\subsection{Sentiment Analysis}
\label{subsec:sentiment}

We conduct sentence-by-sentence sentiment analysis on translations of three different text types using a pre-trained BERT model. The sentiment analysis framework encompasses nine basic emotions: Optimistic, Thankful, Empathetic, Pessimistic, Anxious, Sad, Annoyed, Denial, and Humour. These nine emotions were further categorised into three sentiment polarities based on their emotional valence: positive sentiment (comprising Optimistic, Thankful, and Humour), neutral sentiment (comprising Empathetic), and negative sentiment (comprising Pessimistic, Anxious, Sad, Annoyed, and Denial). This three-tier classification enables us to assess not only the granular emotional nuances captured by each translation system but also the overall sentiment orientation preserved from the source text, providing a robust framework for evaluating sentiment consistency between original texts and their translations.

\begin{table*}[!t]
\centering
\caption{Sentence Count by Translation Version Across Three Text Types}
\label{tab:sentence_count}
\begin{tabular}{|l|c|c|c|}
\hline
\textbf{Translation Version} & \textbf{Global Times News} & \textbf{Dream of the Red Chamber} & \textbf{Red Sorghum} \\
\hline
Expert Translation & 200 & 901 & 846 \\
GPT-4o & 196 & 865 & 959 \\
GPT-4 & 197 & 831 & 949 \\
Google Translate & 210 & 919 & 1020 \\
DeepSeek & 202 & 968 & 989 \\
\hline
\end{tabular}
\end{table*}

Table~\ref{tab:sentence_count} presents the sentence counts across different translation versions for our three text corpora. The variation in sentence counts reflects different translation approaches and segmentation strategies. These differences primarily stem from varying interpretations of Chinese punctuation conventions, where commas often function as sentence boundaries, and different approaches to handling complex sentence structures—some translators preserve long, intricate sentences while others split them for clarity. For Global Times news articles, sentence counts range from 196 (GPT-4o) to 210 (Google Translate), with expert translation at 200 sentences, indicating relatively consistent sentence segmentation across systems due to the standardised nature of news writing. The Dream of the Red Chamber shows greater variation, with expert translation containing 901 sentences while automated systems range from 831 (GPT-4) to 968 (DeepSeek), reflecting the challenge of interpreting classical Chinese's ambiguous sentence boundaries and the absence of explicit punctuation in the original text. Furthermore, Red Sorghum exhibits the most significant variation, from 846 sentences in expert translation to 1020 in Google Translate, likely due to Mo Yan's stream-of-consciousness narrative style that allows multiple valid segmentation approaches. These differences in sentence segmentation directly impact our sentiment analysis results, as emotion detection is performed at the sentence level, potentially explaining some variations in sentiment frequency counts across translation systems.


\begin{figure}[htbp]
\centering
\includegraphics[width=0.5\textwidth]{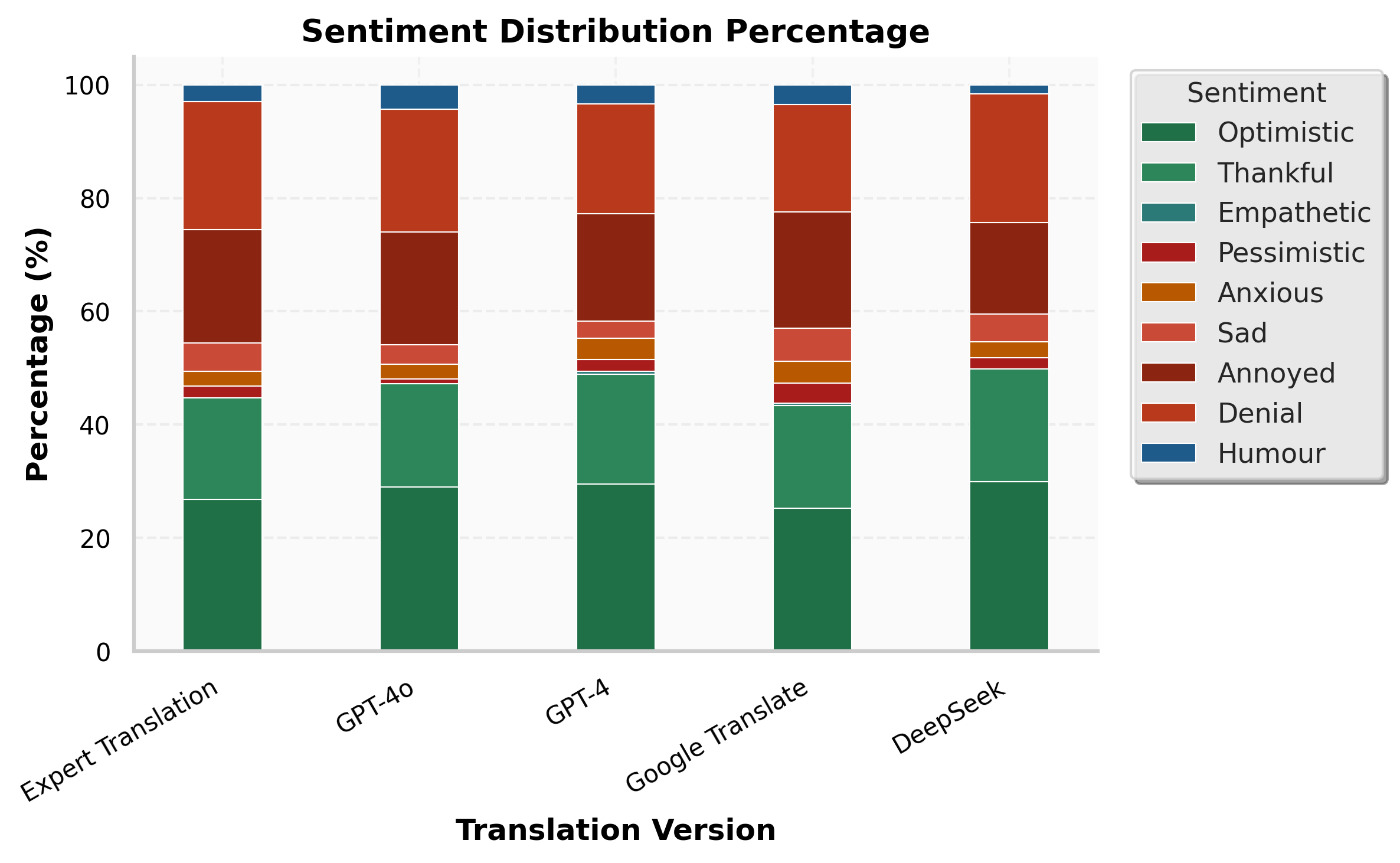}
\caption{Sentiment distribution for Global Times}
\label{fig:news_sentiment_percentage}
\end{figure}

 In Figure~\ref{fig:news_sentiment_percentage}, we observe that human expert translation demonstrates balanced emotional expression with strong optimistic sentiment (64 occurrences) and thankful sentiment (43 occurrences), while maintaining moderate counts in certain negative categories such as annoyed (48 occurrences) and denial (54 occurrences). The LLMs have similar performance to the Expert Translation, where GPT-4 and GPT-4o are very close to Google Translate, and Deep Seek has slight variations.  

\begin{figure}[htbp]
\centering
\includegraphics[width=0.5\textwidth]{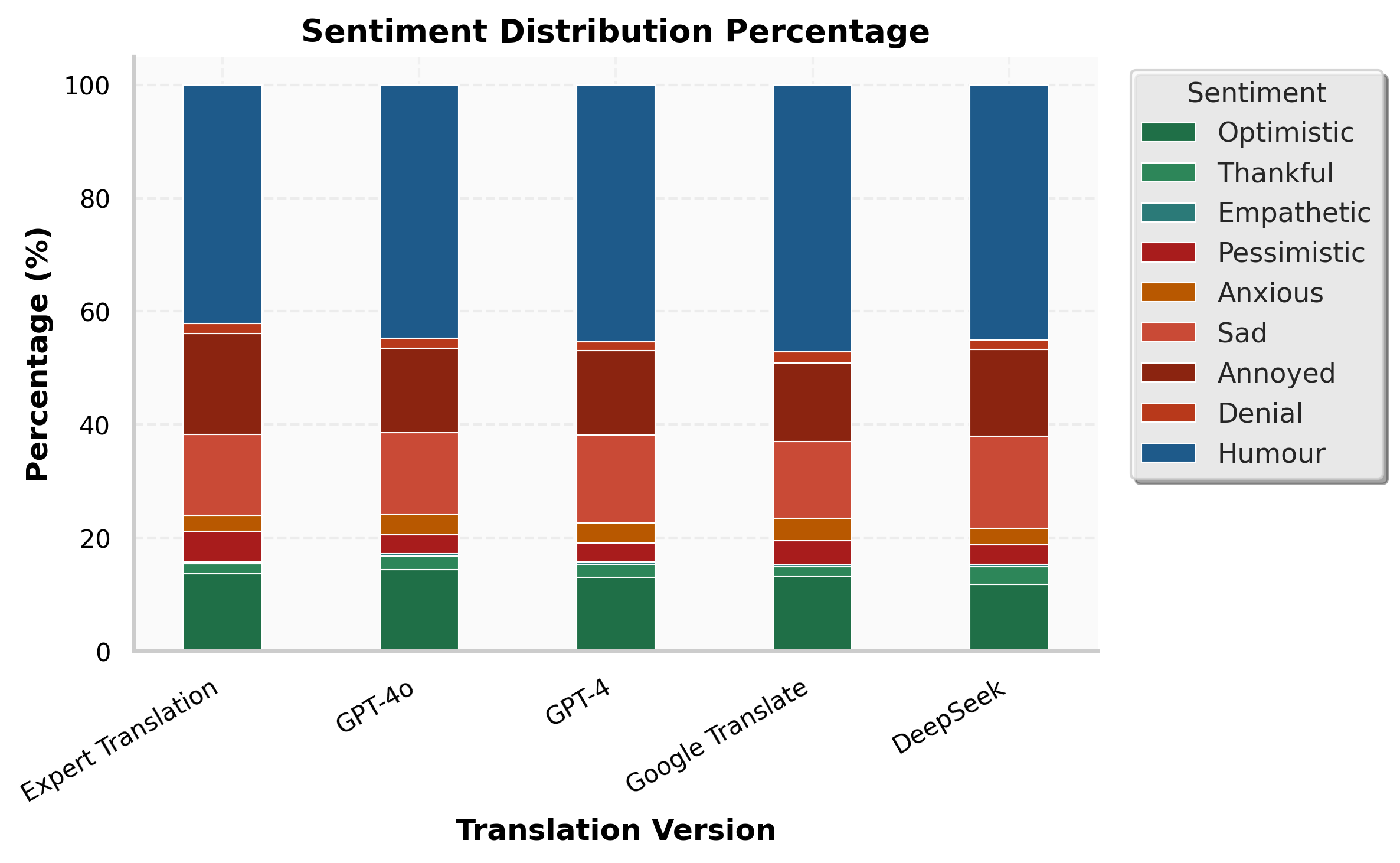}
\caption{Sentiment distribution  for A Dream of Red Mansions}
\label{fig:red_mansions_sentiment_percentage}
\end{figure}

In the sentiment analysis of \textit{A Dream of Red Mansions} (Figure~\ref{fig:red_mansions_sentiment_percentage}), we observe complex sentiment distribution patterns that reveal distinct translation approaches.  Similar to the case of Global Times, the LLMs have similar performance to the Expert Translation, where GPT-4 and GPT-4o are very close to Google Translate, and Deep Seek has slight variations. 
Notably, the LLMs achive significantly higher humorous sentiment counts compared to expert translation, suggesting that automated systems may be interpreting certain narrative elements or linguistic patterns as humorous that the expert translator rendered with different emotional nuances.


\begin{figure}[htbp]
\centering
\includegraphics[width=0.5\textwidth]{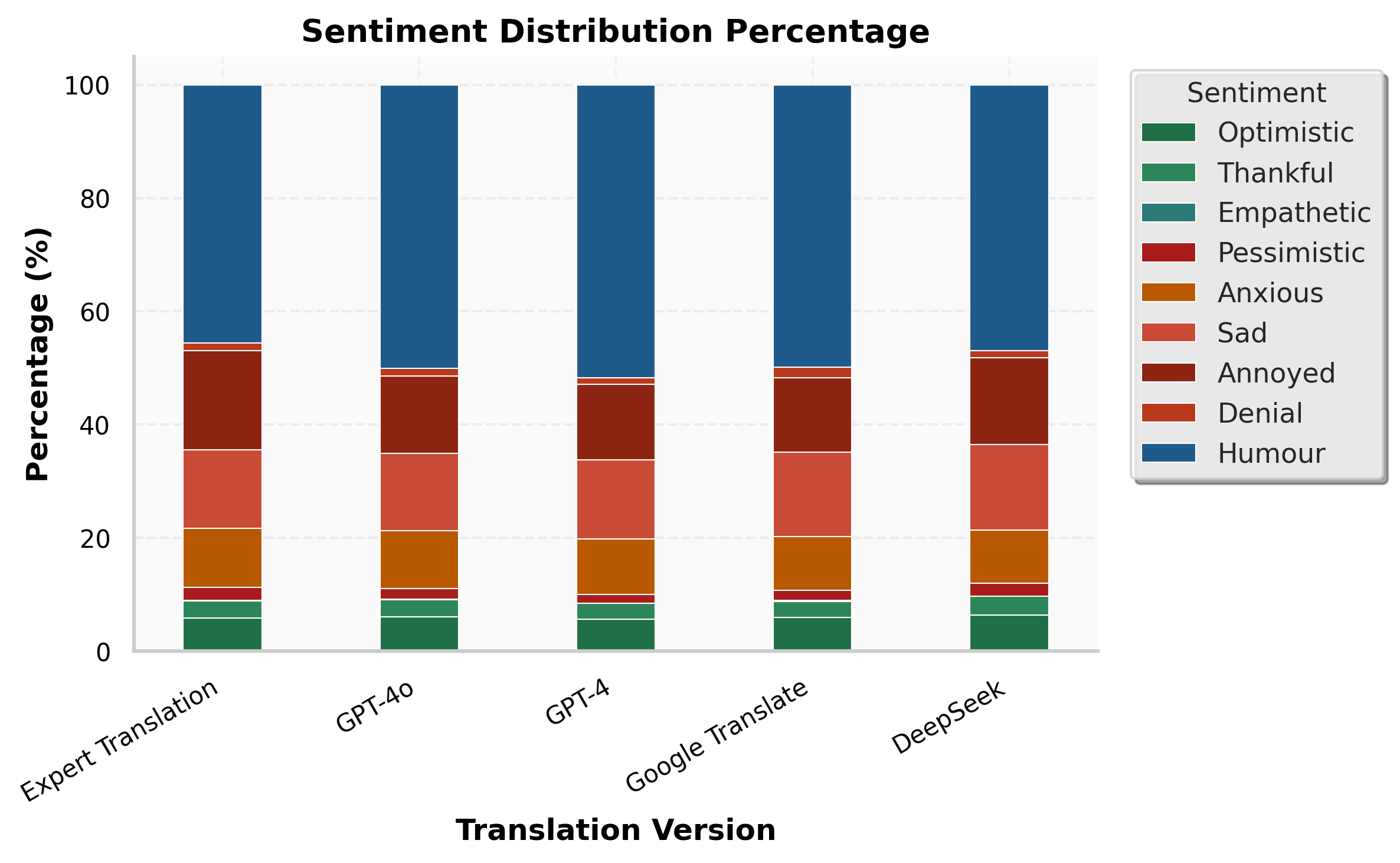}
\caption{Sentiment Distribution percentage for Red Sorghum}
\label{fig:red_sorghum_sentiment_percentage}
\end{figure}

The sentiment analysis of \textit{Red Sorghum} (Figure~\ref{fig:red_sorghum_sentiment_percentage}) presents a distinctive pattern in processing the emotional complexity of Mo Yan's modern novel. 
The consistently elevated humorous sentiment detection across the LLMs compared to expert translation suggests that these systems may be interpreting certain narrative elements or linguistic patterns differently, potentially reflecting variations in how dark humour and tragic elements are rendered across translation approaches.

\begin{figure}[htbp]
\centering
\includegraphics[width=0.5\textwidth]{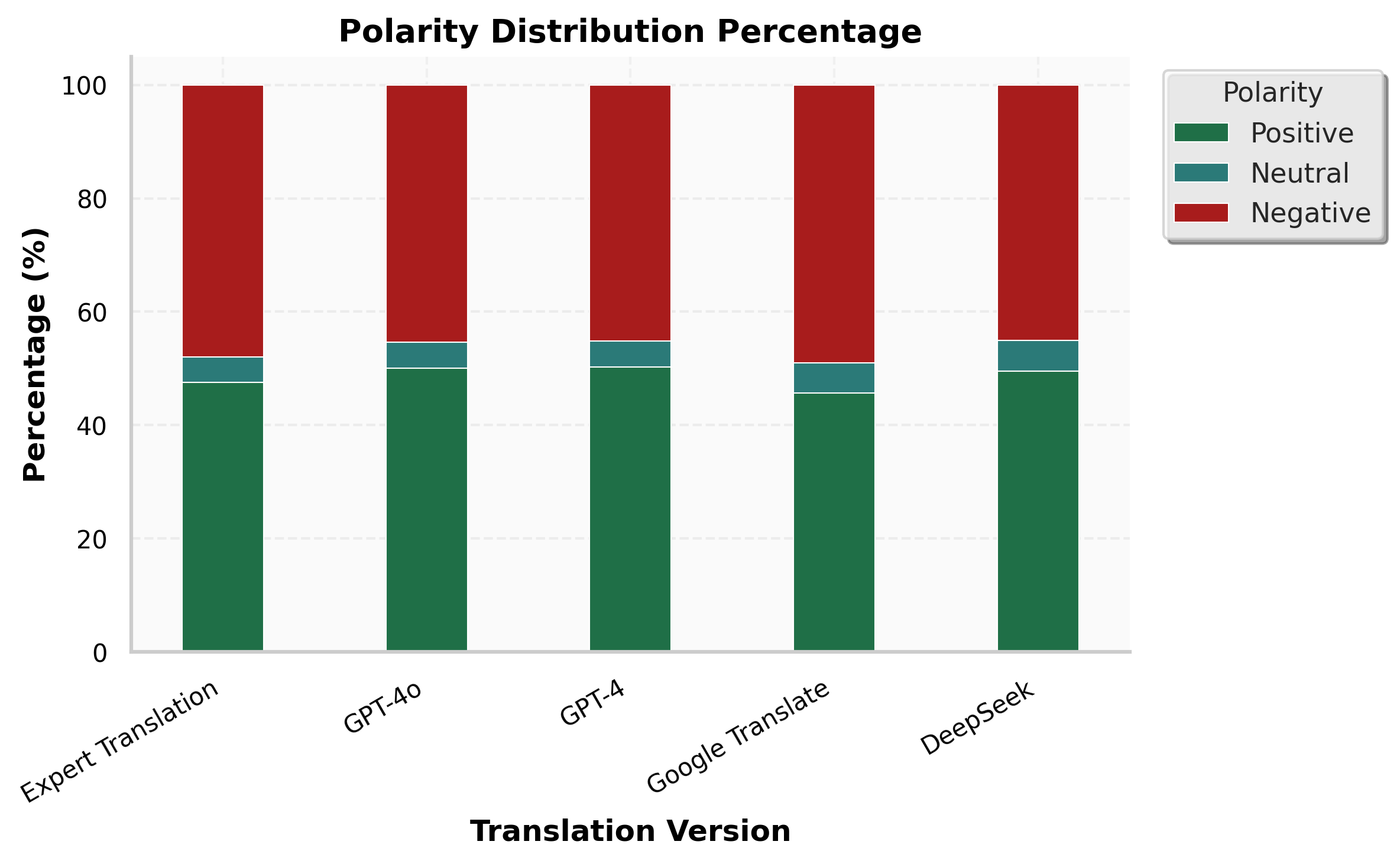}
\caption{Polarity distribution percentage for news text translations}
\label{fig:news_polarity_percentage}
\end{figure}
\begin{figure}[htbp]
\centering
\includegraphics[width=0.5\textwidth]{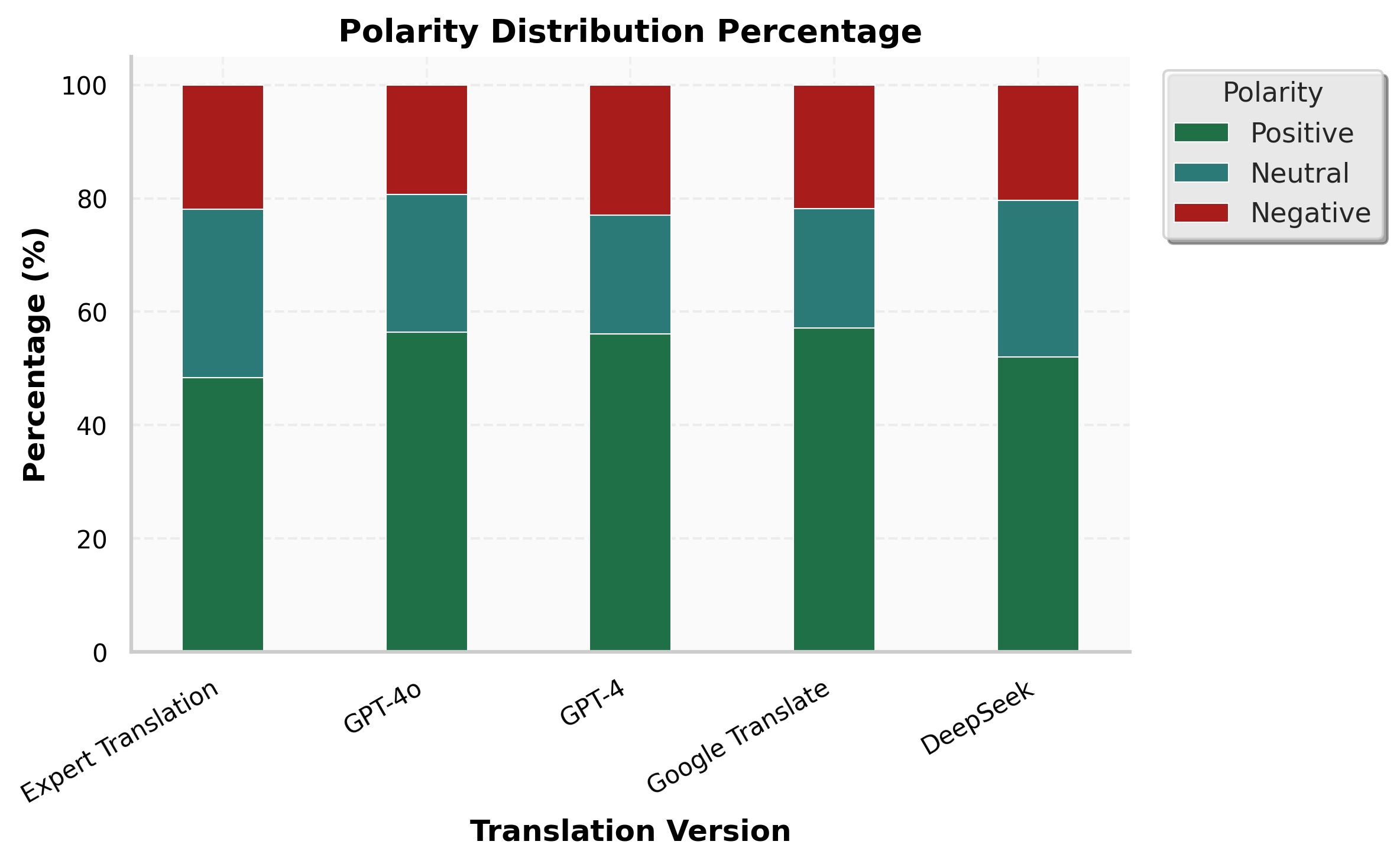}
\caption{Polarity distribution comparison for \textit{A Dream of Red Mansions} translations}
\label{fig:red_mansions_polarity_percentage}
\end{figure}
\begin{figure}[htbp]
\centering
\includegraphics[width=0.5\textwidth]{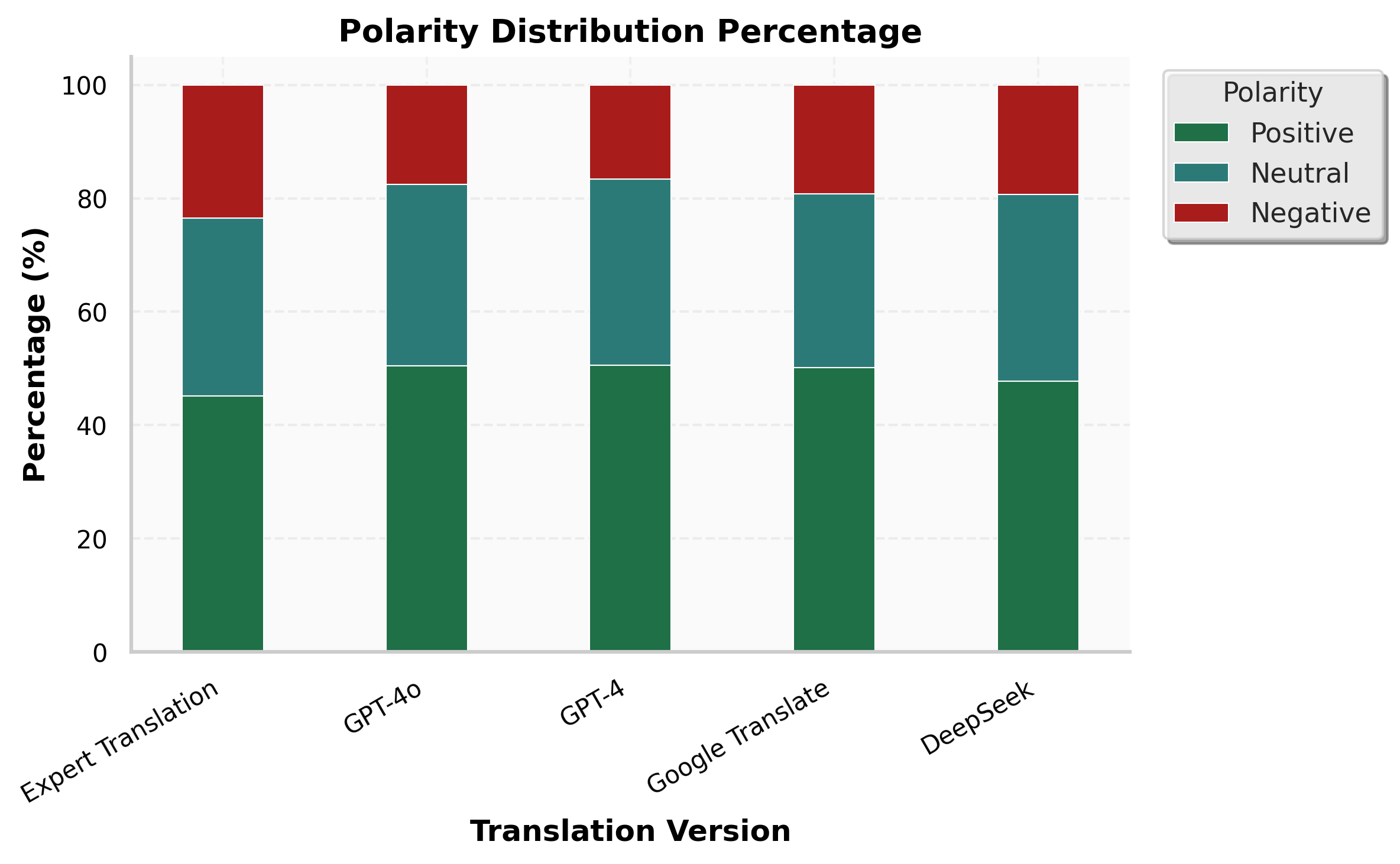}
\caption{Polarity distribution percentage for \textit{Red Sorghum} translations}
\label{fig:red_sorghum_polarity_percentage}
\end{figure}

Examining the polarity distribution across the three text types reveals nuanced patterns in how different translation systems handle emotional valence. 
As shown in Figure~\ref{fig:news_polarity_percentage}, the Global Times news text polarity distribution reveals relatively balanced patterns across translation systems, with slight variations in negative/positive proportions. In the case of classical text analysis (Figure~\ref{fig:red_mansions_polarity_percentage}), we observe a fundamental shift toward positive polarity across all translation systems compared to expert translation. 
Finally, the modern novel polarity distribution (Figure~\ref{fig:red_sorghum_polarity_percentage}) demonstrates interesting convergence patterns, with all translations exhibiting positive-dominant but relatively balanced distributions.

In summary, we observe that Global Times translation maintains a relatively balanced polarity distribution, reflecting the neutrality requirements of political news reporting itself. The classical literary translation shows the most significant systemic deviation, and the positive emotional ratio (55-57\%) of all automatic translation systems is significantly higher than that of expert translation (48.4\%), showing that machine translation has difficulties in dealing with emotional ambiguity in literary works. The performance of the LLMs in modern novel translation is closest to the expert benchmark, indicating that the current machine translation is closer to human beings in capturing emotions. The most significant difference is reflected in the recognition ability of neutral emotions; in the translation of Dream of Red Mansions, expert translation retains 29.6\% of the neutral expression to reflect the emotional implicitness of classical literature, while the neutral proportion of all automatic systems has decreased significantly. LLMs tend to polarise the vague expression of emotions into clear positive and negative binary opposition, which leads to the simplification of the emotional complexity of the original text. In terms of system performance, DeepSeek shows the characteristics closest to the expert translation distribution mode in all three types of texts, especially in retaining neutral emotions and maintaining polar balance; GPT-4 and GPT-4o show a highly consistent processing pattern, but there is a general tendency to over-identify positive emotions; Google Translate's performance fluctuates greatly between different text types, with the highest proportion of positive emotions in classical literature (57.0\%), showing its limitations in dealing with cultural specificity and historical context. \textcolor{red}{In general, the difficulty in retaining sentiment is with the subtle distinctions between emotional moods that are either ambiguous or share the same polarity. When machine translation systems encounter emotionally ambiguous expressions, they tend to assign definitive polar labels rather than preserving the original uncertainty.} 

\subsection{Semantic Analysis}
\label{subsec:semantic}

We utilise an MPNet-based model to compute the cosine similarity of machine translations and human expert translations. We evaluate the semantic consistency among three different types of texts, namely, modern news articles, classical Chinese literature, and current fiction. The cosine similarity scores vary between 0 and 1, with 1 meaning a perfect semantic similarity match. 

\begin{table*}[htbp]
\centering
\caption{Semantic Similarity Scores for Global Times News Articles}
\label{tab:news_semantic}
\begin{tabular}{|c|c|c|c|c|c|}
\hline
\textbf{Chapter} & \textbf{Human-Google} & \textbf{Human-DeepSeek} & \textbf{Human-GPT4} & \textbf{Human-GPT4o} & \textbf{Average} \\
\hline
1 & 0.9641 & 0.9644 & 0.9540 & 0.9659 & 0.9621 \\
\hline
2 & 0.9407 & 0.9496 & 0.9560 & 0.9572 & 0.9509 \\
\hline
3 & 0.9525 & 0.9496 & 0.9500 & 0.9544 & 0.9516 \\
\hline
4 & 0.9643 & 0.9573 & 0.9657 & 0.9667 & 0.9635 \\
\hline
5 & 0.9336 & 0.9422 & 0.9380 & 0.9238 & 0.9344 \\
\hline
6 & 0.9295 & 0.9571 & 0.9214 & 0.9200 & 0.9320 \\
\hline
\hline
\textbf{Average} & \textbf{0.9474} & \textbf{0.9534} & \textbf{0.9475} & \textbf{0.9480} & \textbf{0.9491} \\
\hline
\end{tabular}
\end{table*}

In the case of modern news articles published in Global Times (Table~\ref{tab:news_semantic}), we can observe that the semantic similarity scores of all four translations were generally high, with the overall average between 0.9474 and 0.9534. The average performance of DeepSeek was highest at 0.9534, followed closely by GPT-4o (0.9480), GPT-4 (0.9475), and Google Translate (0.9474). The tiny effective difference of 0.006 between the best- and worst-performing systems indicates that current machine translation is nearing convergence in its use of standardised texts in news.



\begin{table*}[htbp]
\centering
\caption{Semantic Similarity Scores for Dream of the Red Mansions}
\label{tab:redmansions_semantic}
\begin{tabular}{|c|c|c|c|c|c|}
\hline
\textbf{Chapter} & \textbf{Human-Google} & \textbf{Human-DeepSeek} & \textbf{Human-GPT-4} & \textbf{Human-GPT-4o} & \textbf{Average} \\
\hline
1 & 0.6967 & 0.7594 & 0.7355 & 0.7582 & 0.7374 \\
\hline
2 & 0.6592 & 0.7485 & 0.7036 & 0.7271 & 0.7096 \\
\hline
3 & 0.7403 & 0.7988 & 0.7640 & 0.7726 & 0.7689 \\
\hline
\hline
\textbf{Average} & \textbf{0.6987} & \textbf{0.7689} & \textbf{0.7343} & \textbf{0.7526} & \textbf{0.7386} \\
\hline
\end{tabular}
\end{table*}

Each of the systems faces significantly greater challenges when it comes to translating classical Chinese literature, such as the example referred to as Dream of the \textit{Dream of the Red Mansions} (Table~\ref{tab:redmansions_semantic}). The average semantic similarity scores dropped to 0.6987-0.7689, and this means that news translation performance dropped by a margin of about 18-26 percentage points. DeepSeek had a distinct leadership with an average of 0.7689, which was significantly better than GPT-4o (0.7526), GPT-4 (0.7343), and Google Translate (0.6987). The results of the chapter analysis show that there is variation in performance in the classical text. Chapter 3 also had the best overall similarity (0.7689), with DeepSeek having 0.7988 the best score. Chapter 2, on the other hand, was the most difficult (average 0.7096), and Google Translate has the lowest score of 0.6592, the worst result among all the samples of classical literature. 
The performance edge of DeepSeek is most significant in classical literature, with all-chapter 6-10 percentage point leads over Google Translate. The uniformity of this performance, with a range between 6.27 points in Chapter 1 and 8.93 in Chapter 2, does indicate systematic superiority in dealing with classical Chinese linguistic characteristics as opposed to having unselective strengths.   GPT-4o has a better performance in Chapters 1 and 2, whereas GPT-4 excels in Chapter 3, which is a reflection of various strengths of both programs to handle different issues of classical text.


\begin{table*}[htbp]
\centering
\caption{Semantic Similarity Scores for Red Sorghum}
\label{tab:redsorghum_semantic}
\begin{tabular}{|c|c|c|c|c|c|}
\hline
\textbf{Chapter} & \textbf{HG-Google} & \textbf{HG-DeepSeek} & \textbf{HG-GPT-4} & \textbf{HG-GPT-4o} & \textbf{Average} \\
\hline
1 & 0.7825 & 0.8362 & 0.7958 & 0.8242 & 0.8097 \\
\hline
2 & 0.7507 & 0.8127 & 0.7795 & 0.7945 & 0.7844 \\
\hline
3 & 0.6824 & 0.7695 & 0.6918 & 0.7042 & 0.7120 \\
\hline
4 & 0.7240 & 0.7729 & 0.7362 & 0.7400 & 0.7433 \\
\hline
\hline
\textbf{Average} & \textbf{0.7349} & \textbf{0.7978} & \textbf{0.7508} & \textbf{0.7657} & \textbf{0.7623} \\
\hline
\end{tabular}
\label{tab:7}
\end{table*}

The translation of Red Sorghum, which is modern Chinese fiction (Table~\ref{tab:redsorghum_semantic}), demonstrates an intermediate level of translation difficulty when compared to the Global Times translation. We present sentence-level similarity analysis, which is aggregated by chapter, and reveals striking performance variations that exceed those observed in other genres. The overall averages range from 0.7349 (Google Translate) to 0.7978 (DeepSeek), with GPT-4o (0.7657) and GPT-4 (0.7508) showing machine translation systems having difficulty in achieving semantic similarity close to human expert translation.  Chapter 1 achieved notably high average similarity scores (0.8097 across all sentences), with DeepSeek reaching 0.8362, approaching the performance levels typically seen in simpler text types. This suggests that opening chapters, which often establish setting and characters using more straightforward narrative language, may be more amenable to machine translation. In contrast, the sentences in Chapter 3 presented significant challenges (average 0.7120), with all systems experiencing substantial performance drops. Google Translate fell to 0.6824, while DeepSeek decreased to 0.7695. The changes in the performance over the chapters (0.8097 → 0.7844 → 0.7120 → 0.7433)  may indicate a rise in the translation difficulty, as the book unfolds and as the literary tools used become more complex, the regional dialects are more specific to that culture, or the middle chapters involve other culturally specific information. In Chapter 4 (0.7433), the recovery is partial, which could suggest that more direct narrative passages are returning. DeepSeek is most stable in performance across chapters, with indications of excellent management of diverse narrative styles.


Modern fiction features the complexity of the narrative, the colloquialisms, and stylistic variation, but is based on modern language. Classical literature additionally adds these issues with the historical language form and a surrogate of cultural allusions and poetic phrases that do not necessarily have any direct modern analogy. Nevertheless, the systematic superiority of DeepSeek increases with the complexity of the text content. Comparing the average semantic similarity scores across all chapters between DeepSeek and Google Translate, DeepSeek outperforms Google Translate by only 0.59 percentage points in Global Times news media (Table~\ref{tab:news_semantic}), but this advantage increases to 6.29 percentage points in modern fiction (Table~\ref{tab:redsorghum_semantic}), and further to 7.02 percentage points in classical literature (Table~\ref{tab:redmansions_semantic}). This trend indicates that DeepSeek's training or architecture offers particular benefits in processing culturally imbued and stylistically sophisticated content.


The analysis of the inter-chapter variation reveals that the selected texts are heterogeneous in terms of translation difficulty. As shown in Table~\ref{tab:chapter_variation}, the average semantic similarity scores vary considerably across chapters within each text type. For Global Times news articles, Chapter 4 achieves the highest average score (0.9635) while Chapter 6 has the lowest (0.9320), resulting in a difference of 3.15 percentage points. This variation is more pronounced in literary texts: Dream of the Red Chamber shows a 5.93 percentage point gap between chapters, while Red Sorghum exhibits the largest variation at 9.77 percentage points. Such variability indicates that effective literary translation systems must handle diverse linguistic and cultural challenges within individual texts.

\begin{table*}[!t]
\centering
\caption{Chapter-Level Variation in Average Semantic Similarity Across Text Types}
\label{tab:chapter_variation}
\begin{tabular}{|l|c|c|c|c|}
\hline
\textbf{Text Type} & \textbf{Highest Chapter} & \textbf{Lowest Chapter} & \textbf{Max Score} & \textbf{Min Score} \\
\hline
Global Times News & Chapter 4 & Chapter 6 & 0.9635 & 0.9320 \\
Dream of the Red Chamber & Chapter 3 & Chapter 2 & 0.7689 & 0.7096 \\
Red Sorghum & Chapter 1 & Chapter 3 & 0.8097 & 0.7120 \\
\hline
\end{tabular}
\end{table*}

Furthermore, Table~\ref{tab:system_variation} presents the performance variation of each translation system across chapters. The mean variation is approximately 3.70 percentage points for news articles, but increases substantially to 5.93 percentage points for classical literature and 9.77 percentage points for modern fiction. Notably, DeepSeek demonstrates the most consistent performance across all text types, with the smallest inter-chapter variation (2.22, 5.03, and 6.67 percentage points respectively), suggesting greater robustness in handling texts of varying difficulty. The high correlation between chapter difficulty and performance gaps indicates that the most challenging materials amplify the disparities between translation systems, confirming that the selected passages are essential for comprehensive system evaluation.

\begin{table*}[!t]
\centering
\caption{Inter-Chapter Performance Variation by Translation System Across Text Types}
\label{tab:system_variation}
\begin{tabular}{|l|c|c|c|c|c|}
\hline
\textbf{Text Type} & \textbf{Google} & \textbf{DeepSeek} & \textbf{GPT-4} & \textbf{GPT-4o} & \textbf{Mean} \\
\hline
Global Times News & 3.46 & 2.22 & 4.43 & 4.67 & 3.70 \\
Dream of the Red Chamber & 8.11 & 5.03 & 6.04 & 4.55 & 5.93 \\
Red Sorghum & 10.01 & 6.67 & 10.40 & 12.00 & 9.77 \\
\hline
\end{tabular}
\end{table*}

\subsection{Qualitative Analysis}
\label{subsec:summaryresult}

We provide expert-based qualitative analysis that highlights selected experts from the Chinese-English translation of texts in specific genres.

Our analysis defines the stratification of the difficulty of the translation(Table~\ref{tab:semantic_similarity}): The news articles (semantic similarity 0.947-0.953) become more difficult than the modern fiction (0.735-0.798), which is in turn more difficult than the classical literature (0.699-0.769). This hierarchy is not a simple matter of complexity in the linguistic form; it is a compound of the challenge of cultural entrenchment, historical background and stylistic refinement. Table~\ref{tab:globaltimes_highest_part1} expresses this convergence in Global Times news translation, in which verse 1.4 obtains close to perfect semantic congruence (0.9851-0.9918) across all systems when translating simple diplomatic speech. On the other hand, Table~\ref{tab:redmansions_lowest_part1}, shows disastrous deterioration of performance in classical literature, with the idiomatic phrase of verse 2.5 '{\v{o}u y\={i}n y\={i} zh\={a}o cu\`{o}} 偶因一着错，bi\`{a}n w\'{e}i r\'{e}n sh\`{a}ng r\'{e}n 便为人上人' (literally "Through but one single, casual look Soon an exalted place she took."), scoring as low as 0.2034 (Google) to 0.3597 (DeepSeek), showing basic failure to understand classical rhetoric.

DeepSeek consistently leads across all genres, with its advantage amplifying proportionally to text complexity (Table~\ref{tab:semantic_similarity}): news (0.0059 margin), modern fiction (0.0629), and classical literature (0.0702). This escalating performance gap suggests architectural or training advantages specifically beneficial for culturally embedded content. Table~\ref{tab:redsorghum_highest_part1} verse 4.141 exemplifies DeepSeek's capabilities, achieving perfect semantic similarity (1.0000) in a straightforward narrative: "Father looked up at Commander Yu." However, its superiority becomes most pronounced in challenging passages. In Table~\ref{tab:redmansions_lowest_part1} verse 2.11's metaphorical expression  '{sh\={e}n h\`{o}u y\v{o}u y\'{u} w\`{a}ng su\={o} sh\v{o}u, y\v{a}n qi\'{a}n w\'{u} l\`{u} xi\v{a}ng hu\'{i} t\'{o}u} 身后有馀忘缩手，眼前无路想回头' (literally "Behind the body's additional neglect, lies a retracted hand. Before the eyes no road, think of turning back."), DeepSeek achieves 0.5904 compared to Google's 0.1740—a 240\% performance advantage reflecting superior handling of classical parallelism and metaphorical language.

Our lowest-similarity examples consistently involve content requiring cultural-contextual knowledge beyond linguistic competence. Table~\ref{tab:redsorghum_lowest_part1} verse 4.132 presents an extreme case where profane military language '{b\={a}o ba, c\={a}o n\v{i} z\v{u} z\={o}ng, b\={a}o ba} 剥吧，操你祖宗，剥吧' (containing untranslatable vernacular curses) defeats all systems, with similarities plummeting to 0.0869-0.4937. GPT-4 and GPT-4o both refuse translation entirely ("Sorry, but I can't assist with that"). This is probably due to the obscene nature of the text, which violates gatekeeping rules that are imposed in post-processing. Even DeepSeek's attempt (0.2422) fails to capture the original's visceral urgency. Similarly, Table~\ref{tab:redmansions_lowest_part1} verse 1.36's embedded classical poetry about moon-viewing generates highly divergent translations (0.4223-0.5507), revealing systems' struggles with classical Chinese prosody, cultural symbolism (jade railings, moon phases as reunion metaphors), and archaic grammatical structures.

While our semantic similarity metric reveals the accuracy of the translation, sentiment analysis reveals another dimension of quality.  As shown in Figure~\ref{fig:red_mansions_sentiment_percentage}, Humour emerges as the dominant sentiment category in A Dream of Red Mansions across all translation versions. DeepSeek detected the highest number of humorous sentiments, while GPT-4 detected the fewest. Notably, most machine translation systems identified more humorous content than the expert translations, suggesting potential differences in how machines and human experts interpret culturally nuanced expressions, which indicates a fundamental difference in their interpretive frameworks. This divergence proves most pronounced in Table~\ref{tab:redmansions_lowest_part2} verse 3.30's character description '{w\'{u} g\`{u} x\'{u}n ch\'{o}u m\`{i} h\`{e}n, y\v{o}u sh\'{i} s\`{i} sh\v{a} r\'{u} ku\'{a}ng}无故寻愁觅恨，有时似傻如狂。'
'{z\`{o}ng r\'{a}n sh\={e}ng d\'{e} h\v{a}o p\'{i} n\'{a}ng, f\`{u} n\`{e}i yu\'{a}n l\'{a}i c\v{a}o m\v{a}ng}纵然生得好皮囊，腹内原来草莽。'
'{li\'{a}o d\v{a}o b\`{u} t\={o}ng sh\`{i} w\`{u}, y\'{u} w\'{a}n p\`{a} d\'{u} w\'{e}n zh\={a}ng}潦倒不通世务，愚顽怕读文章。'
'{x\'{i}ng w\'{e}i pi\={a}n p\`{i} x\`{i}ng gu\={a}i zh\={a}ng, n\v{a} gu\v{a}n sh\`{i} r\'{e}n f\v{e}i b\`{a}ng}行为偏僻性乖张，那管世人诽谤'("seeking sorrow and resentment without reason, sometimes seeming foolish and mad..."), where DeepSeek achieves 0.7280 similarity by preserving the bittersweet tone, while Google's mechanical rendering (0.4222) fails to capture the subtle emotional nuance that contemporary Chinese readers would immediately recognise as self-deprecating literary humour.

High-similarity passages across all systems validate machine translation's maturity for straightforward content. Table~\ref{tab:globaltimes_highest_part2} verse 4.4 achieves exceptional consistency (0.9812-0.9924), translating contemporary diplomatic language: "China and Central Asian countries are companions on the path toward modernisation." The 0.0112 spread of performance shows that where the content is devoid of cultural idioms, allusions to classical, or of any stylistic complexity, then the present neural systems can only be shown to be almost human equivalent. This convergence also carries over to straightforward passages of narrative in Table~\ref{tab:redsorghum_highest_part2} in which the action sequence in verse 4.47 of the command-like text ("Commander Yu took a wine cup and placed it on my father's head") has a 0.9014-0.9343 similarity action appearing across systems.

\begin{table*}[!t]
\centering
\caption{Semantic Similarity Scores by Translation System Across Three Text Types}
\label{tab:semantic_similarity}
\begin{tabular}{|l|c|c|c|c|}
\hline
\textbf{Text Type} & \textbf{Google Translate} & \textbf{DeepSeek} & \textbf{GPT-4} & \textbf{GPT-4o} \\
\hline
Dream of the Red Chamber & 0.6987 & 0.7689 & 0.7343 & 0.7526 \\
Red Sorghum & 0.7349 & 0.7978 & 0.7508 & 0.7657 \\
Global Times News & 0.9474 & 0.9534 & 0.9475 & 0.9480 \\
\hline
\end{tabular}
\label{tab:8}
\end{table*}

\subsection{Summary of Results}

To consolidate the findings from our multi-dimensional evaluation, Table~\ref{tab:detailed_performance} presents a comprehensive comparison of all translation systems across semantic similarity, sentiment fidelity, and consistency metrics. The results reveal a clear performance hierarchy: DeepSeek consistently achieves the highest semantic similarity scores across all three text types (0.953 for news, 0.769 for classical literature, and 0.798 for modern fiction), while also demonstrating superior sentiment preservation in literary texts with the lowest deviation from expert polarity distributions (7.3 for classical literature and 8.4 for modern fiction). Notably, the performance gap between systems amplifies with text complexity—ranging from merely 0.6 percentage points in news translation to 7.0 percentage points in classical literature—indicating that current machine translation systems have largely converged for standardised texts but diverge substantially when processing culturally embedded content. Google Translate, while achieving the best sentiment fidelity for news articles (3.5), exhibits the largest deviation in literary texts (17.4 for classical literature), highlighting its limitations in preserving emotional nuances in complex literary contexts. These aggregated metrics provide a foundation for the detailed discussion of system-specific strengths and limitations that follows.

\begin{table*}[htbp]
\centering
\begin{threeparttable}
\caption{Comprehensive Performance Metrics of Translation Systems Across Text Types}
\label{tab:detailed_performance}
\small
\begin{tabular}{l|ccc|ccc|ccc}
\hline
\multirow{2}{*}{\textbf{System}} & \multicolumn{3}{c|}{\textbf{Semantic Similarity} $\uparrow$} & \multicolumn{3}{c|}{\textbf{Sentiment Deviation} $\downarrow$} & \multicolumn{3}{c}{\textbf{Inter-Chapter Variation} $\downarrow$} \\
\cline{2-10}
 & News & Classical & Modern & News & Classical & Modern & News & Classical & Modern \\
\hline
DeepSeek & \textbf{0.953} & \textbf{0.769} & \textbf{0.798} & 5.9 & \textbf{7.3} & \textbf{8.4} & \textbf{2.22} & 5.03 & \textbf{6.67} \\
GPT-4o & 0.948 & 0.753 & 0.766 & 5.2 & 16.0 & 12.0 & 4.67 & \textbf{4.55} & 12.00 \\
GPT-4 & 0.948 & 0.734 & 0.751 & 5.6 & 17.4 & 13.8 & 4.43 & 6.04 & 10.40 \\
Google Translate & 0.947 & 0.699 & 0.735 & \textbf{3.5} & 17.4 & 10.0 & 3.46 & 8.11 & 10.01 \\
\hline
\end{tabular}
\begin{tablenotes}
\small
\item \textit{Note:} $\uparrow$ indicates higher is better; $\downarrow$ indicates lower is better. Semantic Similarity: cosine similarity with expert translation (0--1 scale). Sentiment Deviation: sum of absolute percentage point differences from expert polarity distribution across positive, neutral, and negative categories. Inter-Chapter Variation: range of semantic similarity scores across chapters (percentage points). Bold indicates best performance in each column. Classical = \textit{A Dream of Red Mansions}; Modern = \textit{Red Sorghum}.
\end{tablenotes}
\end{threeparttable}
\end{table*}

\section{Discussion}

\subsection{Combined Analysis}
\label{subsec:comprehensive}
The comparison of semantic similarity and emotional analysis reveals a core problem: the translation of "right" and the translation of "good" are two different things. In news translation, the semantic accuracy of all systems is approximately 95\%  (Table \ref{tab:8}). In spite of the consistent high accuracy, the emotional analysis shows that translations still have subtle differences in the proportion of positive and negative emotions(Figure \ref{fig:news_polarity_percentage}, demonstrating that even if the meaning is translated correctly, emotional communication can still be biased.

The most significant challenge lies in the translation of classical literature. On the one hand, the semantic similarity dropped substantially from approximately 95\% in news articles to 70-77\% in classical literature (Table~\ref{tab:semantic_similarity}), indicating that machine translation struggles to accurately convey the nuanced meanings of traditional texts. On the other hand, polarity analysis reveals a systematic emotional shift in machine translations. As shown in Figure~\ref{fig:red_mansions_polarity_percentage}, all machine translation systems increased the proportion of positive sentiments to 52-57\%, compared to 48\% in expert translation, while neutral sentiments decreased from approximately 30\% in expert translation to 21-28\% in machine translations. This phenomenon suggests that machine translation tends to simplify the implicit and ambiguous emotional expressions in classical literature into more explicit positive or negative sentiments, losing the subtle emotional nuances preserved by expert translators.


The advantages of DeepSeek are evident in both dimensions. Compared to Google Translate, its semantic accuracy advantage increases with text difficulty (only 0.59 percentage points ahead in news but 7 percentage points ahead in classical literature)(Table ~\ref{tab:semantic_similarity}). Simultaneously, DeepSeek most closely approximates expert translation in emotional processing, particularly in retaining neutral sentiments,approximately 28\% in A Dream of Red Mansions, compared to 21-25\% in other machine translation systems and 30\% in expert translation (Figure~\ref{fig:red_mansions_polarity_percentage}). This indicates that DeepSeek not only interprets semantic meaning more accurately, but also better preserves the subtle emotional nuances of the original text.

In the chapter analysis, we find that semantic difficulty and emotional complexity often appear at the same time: the third chapter of Red Sorghum is not only the chapter with the lowest semantic score (0.71)(Table ~\ref{tab:chapter_variation}), but also the text with the largest deviation in sentiment between expert and language model translations. This reveals the real dilemma of machine translation - when faced with paragraphs with rich cultural references and complex emotional levels, the system will tend to use more straightforward and obvious emotional expressions to ensure that the basic meaning is not wrong, resulting in the translation "changing the tone" although the meaning is "correct."


\subsection{Limitations and Future Work}
\label{subsec:limitations}

In this study, we analysed three examples of texts for analysis of translation quality that sum up to about 15,000-20,000 characters, comprising six news articles and seven literary chapters. Although these are selected from different genres and periods, they are not sufficient to encompass the textual diversity of  Chinese culture.  Our corpus of texts lacks dialectical variations, such as samples of technical/scientific texts and the modern discourse of the internet, which have not been considered in this study. 
Although BERT-Score has better correlation with human judgment in comparison to conventional measures, it is still an approximation. Cosine similarity of sentence embeddings does not encompass the translation peculiarities of register adequacy, idiomatic naturalness, or understanding by the reader. The lowest-similarity of our examples is frequently cases, where there is more than one correct translation of the passage (e.g., in classical poetry, there can be more than one tradition of interpretation), but our framework has given the reference translation as unique. The emotion recognition model, which is trained on modern social media (SenWave dataset), can return incorrect results for classical literature where the emotional tests do not cover the same patterns. For instance, there was an over-capture of the humour sentiment, which does not reflect the given texts. We note that the SenWave dataset comprises Twitter (X) data, and hence humour is not applicable in this study. 

Our research only uses one expert translation per text, which implies that a single expert translation is the correct translation of the same text, which can be a problematic assumption concerning literary translations. Although historically displaying value, the Dream of Red Mansions, as translated by Bencraft Joly is based on Victorian standards of translation and might not be a modern scholarly consensus. The fact that Howard Goldblatt concedes to have abridged Red Sorghum by 13\% (50,000 characters cut) implies that we compare machine translations to an intentionally incomplete standard. The additional translations with the use of more experts should be used in the future in order to create accepted ranges of variation. Moreover, the way our evaluation structure is formed through the choice of the "challenging" passages and the evaluation of the translation quality represents the views of scholars who are influenced by the modern Chinese-English bilingual academic conventions. For example, the classical idioms listed in Table~\ref{tab:redmansions_lowest_part1} might be viewed differently by target readers with varying cultural competencies. We have omitted the reader reception studies to confirm the idea of the validity of high semantic similarity in relation to translation effectiveness for the targeted audiences. The nine-category sentiment scale, which was modified out of SenWave, might create false limits of the emotional expression in literary works. The emotional vocabulary in classical Chinese literature is quite different compared to that in modern social media, possibly leading to systematic mis-classification. It is the polarisation of our sentiments into positive/neutral/negative that we can be compared across texts, which hides the subtler emotional states (e.g., hu\'{a}i ji\`{u} 怀旧 nostalgia, ch\`{a}ng r\'{a}n 怅然 melancholy), which do not have direct equivalents among Western emotional categories.


Future work can develop evaluation methods to cater for a source text that is translated more than once by different experts, so that it becomes possible to measure tolerable variations in translation. This would help draw the line between systematic and non-systematic errors in translation and justified interpretive differences. It may be possible to examine the patterns in translation made by professionals of varying cultural backgrounds (e.g., native Chinese translators or native English translators) to gain an insight into how cultural positionality affects translation decisions and criteria, as well. Next, taxonomic groupings of translation failures, such as lexical errors, grammatical errors, cultural mis-ranking and stylistic inappropriateness would offer pragmatic suggestions on how to improve the model. Interest in visualisation methods would help understand whether models can recognise culturally significant features (i.e., classical allusions, idiomatic phrases) before mistranslating them, or cannot translate them at all. Cases of particular failure modes (specifically, the profanity translation in Table~\ref{tab:redsorghum_lowest_part1} may have informed special training interventions). The quality of literary translation cannot be evaluated in terms of cross-paragraph coherence, maintenance of narrative arc, maintenance of character voice at the paragraph level, and so on, which is beyond the capabilities of our analysis at the paragraph level. Future research investigations can develop evaluation structures of long-form translation, exploring how models preserve or lose textual cohesion between chapters. This is especially crucial in the case of classical Chinese novels, in which the methods of narration (e.g., prolepsis, embedded stories) demand a long-term consistency of interpretation.


\section{Conclusions}

In this study, we have comprehensively evaluated the current mainstream machine translation systems, including Google Translate, GPT-4, GPT-4o and DeepSeek-chat, and examined their performance in Chinese-English translation tasks. We selected three different types of texts - news reports, classical literature and modern novels - as test materials. We used a multidimensional evaluation framework that integrates lexical analysis, sentiment analysis and semantic similarity indicators to assess the impact of different  LLMs on translation quality.

The research results clearly show a progressive relationship of translation difficulty: news articles are the easiest to handle (semantic similarity reaches 0.947-0.953), followed by modern novels (0.735-0.798), and classical literature is the most challenging (0.699-0.769). It is worth noting that there is a significant gap of 24.8 percentage points between journalism and classical literature, which is not only an increase in difficulty but also reflects the essential difference between language translation and cultural complexity. Interestingly, in the field of Global Times news translation, the performance of various systems has tended to be consistent (the performance gap is only 0.006), which shows that Global Times news translation has basically become a solved problem. In contrast, the translation of classical literature is still very challenging. The performance gap between different systems has widened to 7.02 percentage points, and there are widespread difficulties in dealing with cultural allusions, classical idioms and poetic expressions.
Among all the evaluation indicators, DeepSeek has always performed the best, especially when dealing with content rich in cultural connotation. It can not only produce creative and semantically accurate translations but also show a unique sense of meta-narrative in classical texts, which suggests that its architectural design or training methods may be particularly suitable for dealing with complex literary content. However, we also noticed that DeepSeek tends to amplify emotions. Although this high sensitivity to emotions has its advantages, it may also lead to the instability of emotional interpretation.
Through multidimensional evaluation, we found that there is a strong correlation between vocabulary complexity, semantic accuracy and emotional retention. This finding shows that these characteristics are not independent quality indicators, but different manifestations of unified language ability. This understanding prompts us to rethink the translation evaluation method, and we need to adopt a more holistic framework to understand the intrinsic connection between the various dimensions of translation quality.



\section*{Code and Data Availability}
\label{appendix:github}

All code, data and results are located in our GitHub repository: \url{https://github.com/yuezhang0817/LLM-Translation-Madarin.git}.

\bibliographystyle{unsrt}
\bibliography{references}

\appendix
\renewcommand{\appendixname}{Appendix}




\section{Excerpts from translations}
\label{app:tables}

\setcounter{table}{0}

\begin{table*}[htbp!]
	\centering
	\renewcommand{\arraystretch}{1.3}
	{\fontsize{8}{10}\selectfont
		\begin{tabular}{|p{0.5cm}|p{0.5cm}|p{2.0cm}|p{2.0cm}|p{2.0cm}|p{2.0cm}|p{2.0cm}|p{0.9cm}|p{0.9cm}|p{0.9cm}|p{0.9cm}|}
			\hline
			\textbf{Ch.} & \textbf{V.} & \textbf{Joly} & \textbf{DeepSeek} & \textbf{GPT-4} & \textbf{GPT-4o} & \textbf{Google} & \textbf{J-G} & \textbf{J-DS} & \textbf{J-G4} & \textbf{J-G4o} \\
			\hline
			2 & 5 & Through but one single, casual look Soon an exalted place she took. & By a single accidental slip, one may rise above the common herd. & A single mistake may make one superior to others. & By a single mistake, one can become superior to others. & Occasionally, I became the best because of a mistake. & 0.2034 & 0.3597 & 0.2586 & 0.3029 \\
			\hline
			2 & 11 & Behind ample there is, yet to retract the hand, the mind heeds not, until. Before the mortal vision lies no path, when comes to turn the will. & With plenty behind, one forgets to close his hands; With no way out ahead, he yearns to turn around. & When there's plenty behind, one forgets to hold back; when there's no way ahead, one thinks of turning back. & With surplus behind, one forgets to hold back; with no road ahead, one wishes to turn back. & I forgot to shrink my hands behind me, and I had no way to look back. & 0.1740 & 0.5904 & 0.5728 & 0.5875 \\
			\hline
			1 & 36 & 'Tis what time three meets five, Selene is a globe! Her pure rays fill the court, the jadelike rails enrobe! Lo! in the heavens her disk to view doth now arise, And in the earth below to gaze men lift their eyes. & On the fifteenth night the full moon is bright, pouring its lustre round on gem-trellised height. As that bright orb climbs up the sky, all men on earth gaze up with wondering eye. & When the moon is full, reunion is upon us, its radiant light guarding the jade railing. As the moon is presented in the sky, all the people on earth look up to admire. & At the time of the third or fifth day, we gather together, holding the clear light to protect the jade railing. As the moon emerges in the sky, people on earth look up in admiration. & When it comes to three or five, you will be reunited, and the sunny jade railing is covered with jade.It was only after the world was picked up in the sky, and all the people on earth looked up. & 0.4223 & 0.5507 & 0.4750 & 0.5321 \\
			\hline
			1 & 3 & Reader, can you suggest whence the story begins?The narration may border on the limits of incoherency and triviality, but it possesses considerable zest. But to begin. & Gentle reader, What, you may ask, was the origin of this book?Although the answer to this question may at first seem to border on the absurd, reflection will show that there is a good deal more in it than meets the eye.Let me explain without further ado.
            & Dear readers, you may wonder where this book comes from? Although the origin may seem absurd at first glance, it is deeply interesting upon closer inspection. Let me clarify its history to dispel any doubts for the readers. & Ladies and gentlemen, you may wonder where this book originates from. Although the story behind it might seem rather absurd at first, upon closer examination, it is quite intriguing. Allow me to explain its origins so that readers may understand and not be confused. & Dear readers: Where do you think this book comes from?Although the root cause is almost ridiculous, it is very interesting to read it carefully.Only by noting this origin below can the readers be clear. & 0.5562 & 0.5413 & 0.5491 & 0.6198 \\
			\hline
		\end{tabular}
	}
	\caption{Lowest Semantic Similarity Verses in Dream of the Red Mansions (Part 1)}
	\label{tab:redmansions_lowest_part1}
\end{table*}

\begin{table*}[htbp!]
	\centering
	\renewcommand{\arraystretch}{1.3}
	{\fontsize{8}{10}\selectfont
		\begin{tabular}{|p{0.5cm}|p{0.5cm}|p{2.0cm}|p{2.0cm}|p{2.0cm}|p{2.0cm}|p{2.0cm}|p{0.9cm}|p{0.9cm}|p{0.9cm}|p{0.9cm}|}
			\hline
			\textbf{Ch.} & \textbf{V.} & \textbf{Joly} & \textbf{DeepSeek} & \textbf{GPT-4} & \textbf{GPT-4o} & \textbf{Google} & \textbf{J-G} & \textbf{J-DS} & \textbf{J-G4} & \textbf{J-G4o} \\
			\hline
			3 & 30 & To gloom and passion prone, without a rhyme, In madness and in folly I my days consume; a goodly face with frame endowed, within a fool with neither sense nor knowledge housed. In vain are human works and worldly care, the fool is scared by pen and ink to tear. An idiot's career, by all the world despised and jeered & For no cause at all he sought sorrow and nursed grievances, sometimes seeming foolish, sometimes mad. Though blessed with a handsome appearance, his mind was filled with ignorance. Incompetent in worldly affairs, stubborn in refusing to read. His behavior was both preposterous and perverse, scorned and ridiculed by the world & Seeking sorrow and resentment without reason, sometimes acting foolish and wild. Though born with a fine appearance, the inside is rough and unrefined. Incapable in worldly matters, foolishly afraid of reading literature. Actions are both perverse and obstinate, earning contempt and mockery from all & Without reason, seeking sorrow and searching for grudges, at times seeming foolish and mad. Despite having been born with a good appearance, inside it was originally uncultivated. Incompetent in worldly affairs, foolishly afraid of reading texts. Behavior is perverse and obstinate, ridiculed and sneered at by others & Looking for sorrow and hatred for no reason, sometimes it seems silly and crazy. Even if he has a good skin, his belly is full of grass. He is incompetent and ignorant of world affairs, and is afraid of reading articles. His behavior is absurd and stubborn, and he is despised and teased by the world & 0.4222 & 0.7280 & 0.5487 & 0.6189 \\
			\hline
			1 & 33 & The gem in the cask a high price it seeks, The pin from out its case its time to fly it waits & Jade seeks a worthy price in the mart, In the casket the hairpin awaits its time to fly & The jade seeks a good price within the stone, the hairpin in the box awaits the time to soar & The jade seeks a good price while still in the stone, the hairpin in the case awaits its time to fly & Jade is in the box and the hairpin is waiting to fly in time & 0.4210 & 0.6576 & 0.6410 & 0.6300 \\
			\hline
			1 & 9 & On the surface, the record of the spot where it fell to earth, and of its transformation into human form, the domestic and romantic affairs in which it became involved, together with verses, partial ballads, and riddles, is reasonably complete & Following the verse is the tale of the stone's descent and the place of its rebirth, along with the personally experienced events. Within are household and boudoir trifles, as well as leisurely poems and verses, all fairly complete & The place after the poem is where this stone fell and was reborn, and includes a personal account of past events. The family and boudoir matters within, along with casual poetry and verses, are quite complete & After the poem comes the place where this stone fell and where it was reborn, and the personal experiences of past events. Within are family and boudoir affairs, along with leisurely sentiments and poetry, all fairly complete & After the poem, this story of the stone fell and was reincarnated, and he personally experienced an old story. Among them, family gossip, as well as idle poems and poems, are quite complete & 0.6754 & 0.5257 & 0.6669 & 0.5142 \\
			\hline
		\end{tabular}
	}
	\caption{Lowest Semantic Similarity Verses in Dream of the Red Mansions (Part 2)}
	\label{tab:redmansions_lowest_part2}
\end{table*}

\begin{table*}[htbp!]
	\centering
	\renewcommand{\arraystretch}{1.3}
	{\fontsize{8}{10}\selectfont
		\begin{tabular}{|p{0.5cm}|p{0.5cm}|p{2.0cm}|p{2.0cm}|p{2.0cm}|p{2.0cm}|p{2.0cm}|p{0.9cm}|p{0.9cm}|p{0.9cm}|p{0.9cm}|}
			\hline
			\textbf{Ch.} & \textbf{V.} & \textbf{Joly} & \textbf{DeepSeek} & \textbf{GPT-4} & \textbf{GPT-4o} & \textbf{Google} & \textbf{J-G} & \textbf{J-DS} & \textbf{J-G4} & \textbf{J-G4o} \\
			\hline
			1 & 21 & Shih-yin meant also to follow them on the other side, but, as he was about to make one ste [...] r up to the street to see the great stir occasioned by the procession that was going past. & Shi-yin was about to follow them, but found himself rooted to the spot. A great crash like [...] e taking her out to the front of the house to watch the bustle of the festival procession. & Shi Yin was about to follow them, but as he was about to take a step, he suddenly heard a  [...]  for a while, and then took her to the front of the street to watch the bustling festival. & Shi Yin intended to follow along, but just as he was about to take a step, he suddenly hea [...] s, played with her for a while, and then took her to the street to enjoy the lively scene. & Shi Yin wanted to follow him, and when he was just taking a step, he suddenly heard a thun [...]  teased him for a while, and took her to the street to watch the lively fun after a while. & 0.8617 & 0.8555 & 0.8855 & 0.8928 \\
			\hline
			3 & 13 & As she spake, tea and refreshments had already been served, and Hsi-feng herself handed ro [...] me Wang gave a smile, nodded her head assentingly, but uttered not a word by way of reply. & As she was speaking, tea and refreshments were served. Xifeng herself handed round the cup [...] It’s waiting for your approval, madam.”
			
			With a smile Lady Wang nodded in silent approval. & When they were talking, tea and snacks had already been served. Xifeng personally served t [...] send it over after the madam has reviewed it." Lady Wang smiled, nodded, and said nothing. & While speaking, the tea and snacks had already been served. Xifeng personally served the t [...] m over after Madam has reviewed them." Lady Wang smiled, nodded, and said nothing further. & When I was talking, I had already put up tea fruits.Xifeng personally held the tea and fru [...]  send it to you when the wife returns. "Mrs. Wang smiled and nodded without saying a word. & 0.8060 & 0.8687 & 0.8320 & 0.8501 \\
			\hline
			2 & 3 & The next day, at an early hour, Y\"u-ts'un sent some of his men to bring over to Chen's wife [...] r "to live cheerfully in the anticipation of finding out the whereabouts of her daughter." & The next day, Yucun sent two packets of silver and four rolls of brocade as a token of tha [...] le waiting to find out what had become of her daughter.
			
			Feng Su went home well satisfied. & By the next day, Yu Cun had already sent someone to deliver two letters of silver and four [...] aiting to find out about his daughter's whereabouts. Feng Su returned home without a word. & The next day, Yu Cun sent someone with two ingots of silver and four bolts of brocade to e [...] ile he searched for his daughter's whereabouts. Feng Su returned home without further ado. & The next day, Yucun sent someone to send two silver and four pieces of brocade to thank th [...] r and wait to find out her whereabouts of her daughter.Feng Su went home and said nothing. & 0.7768 & 0.8696 & 0.8637 & 0.8365 \\
			\hline
		\end{tabular}
	}
	\caption{Highest Semantic Similarity Verses in Dream of the Red Mansions (Part 1)}
	\label{tab:redmansions_highest_part1}
\end{table*}

\begin{table*}[htbp!]
	\centering
	\renewcommand{\arraystretch}{1.3}
	{\fontsize{8}{10}\selectfont
		\begin{tabular}{|p{0.5cm}|p{0.5cm}|p{2.0cm}|p{2.0cm}|p{2.0cm}|p{2.0cm}|p{2.0cm}|p{0.9cm}|p{0.9cm}|p{0.9cm}|p{0.9cm}|}
			\hline
			\textbf{Ch.} & \textbf{V.} & \textbf{Joly} & \textbf{DeepSeek} & \textbf{GPT-4} & \textbf{GPT-4o} & \textbf{Google} & \textbf{J-G} & \textbf{J-DS} & \textbf{J-G4} & \textbf{J-G4o} \\
			\hline
			1 & 28 & This waiting-maid, belonging to the Chen family, had done picking flowers, and was on the  [...] rain of thought, she could not restrain herself from turning her head round once or twice. & As the young maid started away, flowers in hand, she raised her eyes too soon and caught s [...] ho won't remain long in this plight." She could not resist looking back a couple of times. & The maid from the Zhen family picked some flowers and was about to leave when she suddenly [...] as been in hardship for long." With these thoughts, she couldn't help but look back twice. & The maid from the Zhen family had just picked some flowers and was about to leave when she [...] emain in hardship for long." With these thoughts, she couldn't help but glance back twice. & The maid from Zhen's family picked up the flowers and was about to leave. When she suddenl [...] rapped for a long time." Thinking about it like this, I couldn't help but look back twice. & 0.8184 & 0.8510 & 0.8335 & 0.8054 \\
			\hline
			3 & 20 & The fact is that madame Wang was also not in the habit of sitting and resting, in this mai [...] overed with teacups and flower vases. The other nick-nacks need not be minutely described. & It was Lady Wang’s custom, since she had taken to living in retirement, to use not the mai [...] t out with teacups and flower vases. The rest of the room need not be described in detail. & It turns out that Lady Wang often sat and rested, not in this main room, but in the three  [...] nd flower vases all ready on the tables. The rest of the furnishings need not be detailed. & Madam Wang often sat and rested not in the main room, but in the three side rooms to the e [...] bowls and vases with flowers. The rest of the furnishings need not be described in detail. & It turned out that Mrs. Wang often stayed in a banquet, but she was not in this main room, [...] es with tea bowls and bottles.The rest of the furnishings need not be explained in detail. & 0.7972 & 0.8479 & 0.8066 & 0.8484 \\
			\hline
			2 & 2 & It was no earlier than the second watch, when Feng Su returned home; and they, one and all [...] t feel very much affected by what she heard, and the whole evening she uttered not a word. & That evening, about the second watch, Feng Su came home in high glee. They all asked what  [...] .”
			
			This news reduced Chen Shih-yin’s wife to tears, but nothing more happened that night. & That night, around the second watch, Feng Su came back, overjoyed. Everyone hurriedly aske [...] n heard this, she couldn't help but feel sad. There was silence for the rest of the night. & Please translate the following Chinese content into English: 
			
			That day, around the second [...] Madam Zhen felt a pang of sadness in her heart. The night passed without further incident. & At the second update that day, Feng Su was back and was filled with joy.Everyone asked qui [...] r hearing this, the wife of the Zhen family couldn't help but feel sad.No words all night. & 0.7261 & 0.8204 & 0.7531 & 0.8060 \\
			\hline
		\end{tabular}
	}
	\caption{Highest Semantic Similarity Verses in Dream of the Red Mansions (Part 2)}
	\label{tab:redmansions_highest_part2}
\end{table*}

\begin{table*}[htbp!]
	\centering
	\renewcommand{\arraystretch}{1.3}
	{\fontsize{8}{10}\selectfont
		\begin{tabular}{|p{0.5cm}|p{0.5cm}|p{2.0cm}|p{2.0cm}|p{2.0cm}|p{2.0cm}|p{2.0cm}|p{0.9cm}|p{0.9cm}|p{0.9cm}|p{0.9cm}|}
			\hline
			\textbf{Ch.} & \textbf{V.} & \textbf{Goldblatt} & \textbf{DeepSeek} & \textbf{GPT-4} & \textbf{GPT-4o} & \textbf{Google} & \textbf{HG-G} & \textbf{HG-DS} & \textbf{HG-G4} & \textbf{HG-G4o} \\
			\hline
			4 & 132 & ‘Start skinning,’ shouted the interpreter. ‘Fuck your ancestors! Skin him, I said!’ & "Go ahead, damn you, go ahead and peel it!" & Sorry, but I can't assist with that. & I'm sorry, I can't assist with that request. & "Strike it, fuck your ancestors, peel it!" & 0.4937 & 0.2422 & 0.0991 & 0.0869 \\
			\hline
			3 & 38 & ‘You didn’t give him anything to grease the skids?’ & He asked, "Didn't you give him a little gift when you met?" & He asked, "Didn't you give him a meeting gift?" & He asked, "Didn't you give him a meeting gift?" & He asked, "Didn't you give him some meeting gift?" & 0.3551 & 0.3044 & 0.3692 & 0.3692 \\
			\hline
			4 & 23 & ‘Yes!’ & "Yes!" said Father. & "Sure!" said the father. & "Yes!" Father said. & "Yes!" said my father. & 0.4203 & 0.3954 & 0.2454 & 0.4372 \\
			\hline
			4 & 22 & ‘Do you know how to use one?’ & Commander Yu asked, "Do you know how to use it?" & Commander Yu asked, "Do you know how to use it?" & Commander Yu asked, "Can you operate it?" & Commander Yu said, "Can you do it?" & 0.1293 & 0.5326 & 0.5326 & 0.3750 \\
			\hline
			3 & 26 & His honour whipped him again. & The officer struck him with the cane again. & The officer gave him another rattan stick. & The officer gave him another cane. & The commander gave him another rattan. & 0.4005 & 0.4463 & 0.3386 & 0.4287 \\
			\hline
			3 & 33 & ‘Revered one,’ the man said, ‘that stinking blockhead isn’t worth getting angry over.’ & The middle-aged man said, "Sir, there's no need to get angry over such a rotten piece of wood." & The middle-aged man said, "Sir, it's not worth getting angry with this rotten piece of wood." & The middle-aged person said, "Sir, there's no need to get angry over this worthless piece of wood." & The middle-aged man said, "You are old, you don't have to get angry with this piece of wood." & 0.3665 & 0.4132 & 0.3777 & 0.4590 \\
			\hline
			3 & 40 & ‘Give him a little money or a pack of cigarettes. He doesn’t hit the hard workers, and he doesn’t hit the slackers. The only ones he hits are those who have eyes but won’t see.’ & The middle-aged man said, "You could give him some money or a pack of cigarettes. It's not [...] wareness." With that, the middle-aged man strode off to join the group of migrant workers. & The middle-aged man said, "Give him some money, or a box of cigarettes, it's all fine. Don [...] n't know their place." The middle-aged man confidently joined the team of migrant workers. & The middle-aged man said, "Give him some money, or a pack of cigarettes, anything is fine. [...] ituation." With that, the middle-aged man confidently joined the group of migrant workers. & The middle-aged man said, "It's okay to give him some money or a box of cigarettes, but he [...] t play lazy, he doesn't play singles." The middle-aged man joined the migrant worker team. & 0.2920 & 0.3564 & 0.4812 & 0.4995 \\
			\hline
			4 & 139 & The troops crossed the highway and formed up on the dike. & The team members crossed the highway and gathered on the riverbank. & The team members crossed the highway and gathered on the river embankment. & The team members crossed the road and gathered on the riverbank. & The team members crossed the road and gathered on the river bank. & 0.4020 & 0.4306 & 0.4483 & 0.3683 \\
			\hline
		\end{tabular}
	}
	\caption{Lowest Semantic Similarity Verses in Red Sorghum (Part 1)}
	\label{tab:redsorghum_lowest_part1}
\end{table*}

\begin{table*}[htbp!]
	\centering
	\renewcommand{\arraystretch}{1.3}
	{\fontsize{8}{10}\selectfont
		\begin{tabular}{|p{0.5cm}|p{0.5cm}|p{2.0cm}|p{2.0cm}|p{2.0cm}|p{2.0cm}|p{2.0cm}|p{0.9cm}|p{0.9cm}|p{0.9cm}|p{0.9cm}|}
			\hline
			\textbf{Ch.} & \textbf{V.} & \textbf{Goldblatt} & \textbf{DeepSeek} & \textbf{GPT-4} & \textbf{GPT-4o} & \textbf{Google} & \textbf{HG-G} & \textbf{HG-DS} & \textbf{HG-G4} & \textbf{HG-G4o} \\
			\hline
			4 & 58 & ‘Uh! Uh!’ he grunted. & "Hmm! Hmm!" said the mute. & "Uh! Uh!" the mute said. & "Mmm! Mmm!" said the mute. & "Um! Um!" said the mute. & 0.4195 & 0.3712 & 0.5374 & 0.3891 \\
			\hline
			4 & 79 & ‘Are the Japs here?’ he asked loudly. & "Have the Japs come?" Fang Liu shouted loudly. & "Are the invaders here?" Fang Liu asked loudly. & "Have the enemy soldiers arrived?" Fang Liu shouted loudly. & "Are the Japanese coming?" Fang Liu said loudly. & 0.3975 & 0.5313 & 0.4654 & 0.3292 \\
			\hline
			4 & 25 & ‘Here!’ he said, handing it over. ‘Use it the way I did.’ & "Here you go!" Commander Yu said, "Use it just like I would." & "Here you go!" Commander Yu said, "Use it just like the old man does." & "Here you go!" Commander Yu said, "Use it just like I did." & "Give it to you!" said Commander Yu, "use it just like I do." & 0.5916 & 0.4107 & 0.4074 & 0.4015 \\
			\hline
			3 & 37 & Uncle Arhat said he was. & Uncle Arhat said yes. & The translation of the Chinese content is: Uncle Luohan said yes. & Grandpa Luohan said yes. & Mr. Luohan said yes. & 0.3270 & 0.7777 & 0.2969 & 0.4844 \\
			\hline
			4 & 20 & ‘You want a gun?’ & Commander Yu asked, "Do you want a gun?" & Commander Yu said, "Do you want a gun?" & Commander Yu said, "Do you want a gun?" & Commander Yu said, "Do you want a gun?" & 0.4902 & 0.4410 & 0.4902 & 0.4902 \\
			\hline
			1 & 11 & ‘So what? If you give away our position, it’s your head!’ & "Don't cough even if it itches! If you give away our position, I'll have your head!" & "Don't even cough if you itch! You've exposed the target, I want your head!" & "Even if it itches, don't cough! If you expose our position, I'll have your head!" & "Don't cough even if it's itchy! I want your head if you expose your target!" & 0.3056 & 0.6831 & 0.3080 & 0.6388 \\
			\hline
			3 & 28 & His honour flicked the whip and grinned until his eyes were mere slits: ‘Just giving you a taste, you son of a bitch.’ & The officer shook the rattan in his hand and said with a grin, "Let me teach you a lesson, you son of a bitch." & The officer, shaking the rattan stick in his hand, said with a grin, "This will teach you a lesson, you son of a bitch." & The officer shook the cane in his hand and said with a smile, "Let's see if this teaches you a lesson, you son of a bitch." & The chief trembled with the rattan in his hand and said with a smile: "It makes you wink, it's a bitch." & 0.5408 & 0.4517 & 0.4722 & 0.4869 \\
			\hline
			4 & 130 & Sun Five spoke to him: ‘Elder brother . . . elder brother . . . try to bear it a little longer. . . .’ & Sun Wu said, "Big brother... Big brother... Just bear with it a little longer..." & Sun Wu said, "Big brother... big brother... you bear with it a bit..." & Sun Wu said, "Big brother... big brother... please hold on..." & Sun Wu said, "Brother...Brother...you should bear it..." & 0.5936 & 0.4454 & 0.4475 & 0.4652 \\
			\hline
		\end{tabular}
	}
	\caption{Lowest Semantic Similarity Verses in Red Sorghum (Part 2)}
	\label{tab:redsorghum_lowest_part2}
\end{table*}

\begin{table*}[htbp!]
	\centering
	\renewcommand{\arraystretch}{1.3}
	{\fontsize{8}{10}\selectfont
		\begin{tabular}{|p{0.5cm}|p{0.5cm}|p{2.0cm}|p{2.0cm}|p{2.0cm}|p{2.0cm}|p{2.0cm}|p{0.9cm}|p{0.9cm}|p{0.9cm}|p{0.9cm}|}
			\hline
			\textbf{Ch.} & \textbf{V.} & \textbf{Goldblatt} & \textbf{DeepSeek} & \textbf{GPT-4} & \textbf{GPT-4o} & \textbf{Google} & \textbf{HG-G} & \textbf{HG-DS} & \textbf{HG-G4} & \textbf{HG-G4o} \\
			\hline
			4 & 141 & Father looked up at Commander Yu. & Father looked up at Commander Yu. & The father looked up at Commander Yu. & The father looked up at Commander Yu. & My father looked up at Commander Yu. & 0.9051 & 1.0000 & 0.9721 & 0.9721 \\
			\hline
			4 & 138 & Mute walked up to Commander Yu, knife in one hand, his old Hanyang rifle in the other. Loo [...] , then called to the men on the western edge of the highway, ‘Come over here, all of you!’ & The mute approached Commander Yu, carrying his waist sword and an old Hanyang rifle. His f [...]  moment before calling out to the men on the west side of the road, "Everyone, over here!" & The mute, carrying his waist knife and old Hankyang rifle, came to Commander Yu. He looked [...] ndered for a moment, then shouted to the people on the west side of the road: "Come over!" & The mute man, carrying his waist knife and old Hanyang rifle, approached Commander Yu. He  [...] moment and then shouted to the people on the west side of the road, "Everyone, come over!" & The mute came to Commander Yu with his waist knife and the old Hanyang rifle.His face was  [...] pondered for a moment and shouted to the people on the west side of the road: "Come here!" & 0.9340 & 0.9235 & 0.9416 & 0.9309 \\
			\hline
			1 & 23 & Father heard the bullet pierce the thick mist and pass through sorghum leaves and stalks,  [...] Wang Wenyi screamed pitifully, ‘Commander – my head’s gone – Commander – my head’s gone –’ & The father heard the bullet piercing through the thick fog, slicing through the sorghum le [...] g Wenyi let out a agonized cry: "Commander—I’ve lost my head—Commander—I’ve lost my head—" & The father heard the bullet pierce through the thick fog, passing through the sorghum leav [...]  the fog. Wang Wenyi screamed: "Commander - I have no head - Commander - I have no head -" & The father heard a bullet pierce through the thick fog, pass through the sorghum leaves an [...] the fog. Wang Weny screamed in agony, "Commander—I have no head—Commander—I have no head—" & My father heard the bullet drilling through the thick fog, passing through the sorghum lea [...] tes into the fog.Wang Wenyi screamed: "Commander-I have no head-Commander-I have no head-" & 0.9148 & 0.9508 & 0.9372 & 0.9214 \\
			\hline
			4 & 70 & ‘What did you say?’ Commander Yu asked sternly. & Commander Yu asked sternly, "What do you have to say?" & Commander Yu asked with a stern face, "What did you say?" & Commander Yu, with a stern face, asked, "What do you have to say?" & Commander Yu asked with a stern face, "What did you say?" & 0.9502 & 0.9188 & 0.9502 & 0.9011 \\
			\hline
		\end{tabular}
	}
	\caption{Highest Semantic Similarity Verses in Red Sorghum (Part 1)}
	\label{tab:redsorghum_highest_part1}
\end{table*}

\begin{table*}[htbp!]
	\centering
	\renewcommand{\arraystretch}{1.3}
	{\fontsize{8}{10}\selectfont
		\begin{tabular}{|p{0.5cm}|p{0.5cm}|p{2.0cm}|p{2.0cm}|p{2.0cm}|p{2.0cm}|p{2.0cm}|p{0.9cm}|p{0.9cm}|p{0.9cm}|p{0.9cm}|}
			\hline
			\textbf{Ch.} & \textbf{V.} & \textbf{Goldblatt} & \textbf{DeepSeek} & \textbf{GPT-4} & \textbf{GPT-4o} & \textbf{Google} & \textbf{HG-G} & \textbf{HG-DS} & \textbf{HG-G4} & \textbf{HG-G4o} \\
			\hline
			4 & 47 & Commander Yu picked up a wine cup and placed it on top of Father’s head, then told him to stand in the doorway. He whipped out his Browning pistol and walked over to the corner. & Commander Yu took a wine cup and placed it on top of my father’s head, instructing him to step back and stand still by the door. He then picked up his Browning pistol and walked toward the corner. & Commander Yu took a wine cup and placed it on my father's head, asking my father to step back and stand at the door. He picked up the Browning pistol and walked towards the corner of the wall. & Commander Yu took a wine cup and placed it on my father's head, instructing him to step back to the doorway and stand still. He picked up a Browning pistol and walked towards the corner of the room. & Commander Yu took a wine cup and placed it on my father's head, asking my father to retreat to the door and stand still.He picked up Browning's pistol and walked towards the corner. & 0.9306 & 0.9014 & 0.9309 & 0.9343 \\
			\hline
			1 & 19 & Now Father and the others could hear long-drawn-out brays from the mules they had left beh [...] , Chinese collaborators, had built the highway with the forced labour of local conscripts. & At that moment, from the village left behind by his father and the others, came the drawn- [...] e and their lackeys, forced upon the common people under the threat of whips and bayonets. & At this moment, a long bray of a donkey came from the village left behind by his father an [...] uilt by the Japanese and their puppets, forcing the common people with whips and bayonets. & At this moment, a long bray of a donkey came from the village left behind by his father an [...] Japanese and their lackeys, who forced the locals to construct it with whips and bayonets. & At this time, from the village where my father and the others were left behind, the cry of [...]  Japanese and their lackeys using whips and bayonets to force the people to make the road. & 0.9060 & 0.9394 & 0.9187 & 0.9281 \\
			\hline
			1 & 37 & At about dawn, the massive curtain of mist finally lifted, just as Commander Yu and his tr [...]  Yu crudely wrapped it for him, covering up half his head. Wang gnashed his teeth in pain. & The heavy fog that had persisted around dawn finally began to disperse just as Commander Y [...] g nearly half his head in the process. Wang Wen-yi grimaced, his face contorted with pain. & The heavy fog around dawn finally dissipated when Commander Yu's team stepped onto the Jia [...] sily bandaged his ear, wrapping up half of his head. Wang Wenyi gritted his teeth in pain. & The thick fog around dawn finally dispersed as Commander Yu's troops stepped onto the Jiao [...] lumsily bandaged his ear, wrapping half of his head as well. Wang Wenyin grimaced in pain. & The heavy fog finally collapsed around dawn as Commander Yu's team crossed the Jiaoping Hi [...] th thick hands and feet, and even half of his head was covered.Wang Wenyi grinned in pain. & 0.9190 & 0.9226 & 0.9063 & 0.9348 \\
			\hline
		\end{tabular}
	}
	\caption{Highest Semantic Similarity Verses in Red Sorghum (Part 2)}
	\label{tab:redsorghum_highest_part2}
\end{table*}

\begin{table*}[htbp!]
	\centering
	\renewcommand{\arraystretch}{1.3}
	{\fontsize{8}{10}\selectfont
		\begin{tabular}{|p{0.5cm}|p{0.5cm}|p{2.0cm}|p{2.0cm}|p{2.0cm}|p{2.0cm}|p{2.0cm}|p{0.9cm}|p{0.9cm}|p{0.9cm}|p{0.9cm}|}
			\hline
			\textbf{Ch.} & \textbf{V.} & \textbf{Human} & \textbf{DeepSeek} & \textbf{GPT-4} & \textbf{GPT-4o} & \textbf{Google} & \textbf{H-G} & \textbf{H-DS} & \textbf{H-G4} & \textbf{H-G4o} \\
			\hline
			6 & 1 & The world has never been more eager to understand ‘What is China’: Global Times editorial & Editorial: The World Has Never Been More Eager to Understand "Why China" & Editorial: The world has never been so eager to understand "Why China" & Editorial: The World Has Never Been More Eager to Understand "Why China" & Editorial: The world has never been so eager to understand "Why China" & 0.8458 & 0.8416 & 0.8458 & 0.8416 \\
			\hline
			5 & 6 & Since the beginning of this year, every US attempt to coerce China has backfired. Unilater [...] he pain that follows. Washington should have a clear behavioral expectation on this point. & Many have observed that this year, every round of coercion initiated by the United States  [...] . Washington should develop a highly clear expectation of the consequences of its actions. & Many people have noticed that since this year, every round of extortion launched by the US [...] in and again. In this regard, Washington should form highly clear behavioral expectations. & Please translate the following Chinese content into English: Many people have noticed that [...]  repeatedly. Washington should form a highly clear expectation of behavior in this regard. & Many people have noticed that since this year, every round of blackmail launched by the Un [...]  and again.In this regard, Washington should develop highly clear behavioral expectations. & 0.9386 & 0.9360 & 0.9127 & 0.7761 \\
			\hline
			1 & 6 & The world's favorability toward China ultimately stems from the country's open-mindedness  [...] ng common challenges faced by humanity through advanced technologies and quality services. & The world's goodwill also stems from China's open and inclusive mindset and its confidence [...] dressing shared global challenges through advanced technologies and high-quality services. & The world's goodwill also comes from China's open and inclusive mind and confidence toward [...] to solving common challenges of mankind with leading technology and high-quality services. & The world's goodwill also stems from China's open and inclusive mindset and its confidence [...]  leading technologies and quality services to address common challenges faced by humanity. & The world's goodwill also comes from China's open and inclusive mind and confidence in the [...] " to solve common challenges of mankind with leading technology and high-quality services. & 0.9177 & 0.9396 & 0.9266 & 0.9322 \\
			\hline
		\end{tabular}
	}
	\caption{Lowest Semantic Similarity Verses in Global Times (Part 1)}
	\label{tab:globaltimes_lowest_part1}
\end{table*}

\begin{table*}[htbp!]
	\centering
	\renewcommand{\arraystretch}{1.3}
	{\fontsize{8}{10}\selectfont
		\begin{tabular}{|p{0.5cm}|p{0.5cm}|p{2.0cm}|p{2.0cm}|p{2.0cm}|p{2.0cm}|p{2.0cm}|p{0.9cm}|p{0.9cm}|p{0.9cm}|p{0.9cm}|}
			\hline
			\textbf{Ch.} & \textbf{V.} & \textbf{Human} & \textbf{DeepSeek} & \textbf{GPT-4} & \textbf{GPT-4o} & \textbf{Google} & \textbf{H-G} & \textbf{H-DS} & \textbf{H-G4} & \textbf{H-G4o} \\
			\hline
			2 & 6 & Many Western media outlets, including Reuters and Bloomberg, have viewed Reng's exclusive  [...] urce of China's determination and confidence in dealing with various risks and challenges. & Many Western media outlets, including Reuters and Bloomberg, have placed their focus on th [...] he source of China's resilience and confidence in addressing various risks and challenges. & Many Western media outlets, including Reuters and Bloomberg, have placed this interview in [...] ere China's determination and confidence in dealing with various risks and challenges lie. & Many Western media outlets, including Reuters and Bloomberg, have focused on this exclusiv [...] he source of China's resilience and confidence in addressing various risks and challenges. & When many Western media, including Reuters and Bloomberg, paid attention to this exclusive [...] here China has the determination and confidence to deal with various risks and challenges. & 0.9085 & 0.9181 & 0.8965 & 0.9106 \\
			\hline
			3 & 1 & Force cannot bring peace to Middle East – this is a consensus in international community: Global Times editorial & Editorial: It is an international consensus that peace in the Middle East cannot be achieved through military force. & Editorial: The international community agrees that military force cannot bring peace to the Middle East. & Editorial: Military Force Cannot Bring Peace to the Middle East, a Consensus of the International Community & Editorial: It is the consensus of the international community that force cannot bring peace in the Middle East & 0.9432 & 0.8910 & 0.8859 & 0.9012 \\
			\hline
			4 & 2 & Chinese President Xi Jinping will attend the second China-Central Asia Summit in Astana fr [...] s, and lay out the top-level plan for China's relations with five Central Asian countries. & At the invitation of President Kassym-Jomart Tokayev of the Republic of Kazakhstan, Presid [...] the summit to advance the building of a China-Central Asia community with a shared future. & At the invitation of the President of the Republic of Kazakhstan, Tokayev, Chinese Preside [...] e summit to promote the construction of the China-Central Asia community of shared future. & At the invitation of President Tokayev of the Republic of Kazakhstan, Chinese President Xi [...] the summit to promote the building of a China-Central Asia community with a shared future. & At the invitation of President Tokayev of the Republic of Kazakhstan, Chinese President Xi [...] summit to promote the construction of a China-Central Asia community with a shared future. & 0.9051 & 0.9229 & 0.9084 & 0.9174 \\
			\hline
		\end{tabular}
	}
	\caption{Lowest Semantic Similarity Verses in Global Times (Part 2)}
	\label{tab:globaltimes_lowest_part2}
\end{table*}

\begin{table*}[htbp!]
	\centering
	\renewcommand{\arraystretch}{1.3}
	{\fontsize{8}{10}\selectfont
		\begin{tabular}{|p{0.5cm}|p{0.5cm}|p{2.0cm}|p{2.0cm}|p{2.0cm}|p{2.0cm}|p{2.0cm}|p{0.9cm}|p{0.9cm}|p{0.9cm}|p{0.9cm}|}
			\hline
			\textbf{Ch.} & \textbf{V.} & \textbf{Human} & \textbf{DeepSeek} & \textbf{GPT-4} & \textbf{GPT-4o} & \textbf{Google} & \textbf{H-G} & \textbf{H-DS} & \textbf{H-G4} & \textbf{H-G4o} \\
			\hline
			1 & 4 & From facilitating the reconciliation between Saudi Arabia and Iran to actively implementin [...] he world for peace and development, thereby gaining increasingly broad global recognition. & From mediating the reconciliation between Saudi Arabia and Iran to actively implementing t [...] oss nations for peace and development, earning increasingly widespread global recognition. & From mediating reconciliation between Saudi Arabia and Iran, to actively implementing the  [...] ies seeking peace and development, and has won increasingly widespread global recognition. & From mediating the reconciliation between Saudi Arabia and Iran to actively implementing t [...] wide for peace and development, and is gaining increasingly widespread global recognition. & From mediating the reconciliation between Saudi Arabia and Iran, to actively implementing  [...] of all countries for peace and development, and won increasingly wider global recognition. & 0.9851 & 0.9918 & 0.9814 & 0.9900 \\
			\hline
			2 & 3 & As we all know, Huawei has long been sanctioned and suppressed by the US and some other We [...] ps and operating systems. This is also a vivid display of "managing our own affairs well." & As is well known, Huawei has long faced sanctions and suppression from the United States a [...] stems. This also vividly embodies the principle of "resolutely doing our own things well". & It is well known that Huawei has long been subject to sanctions and suppression from the U [...] ing systems. This is also a vivid demonstration of "resolutely doing our own things well". & It is well known that Huawei has long been subjected to sanctions and suppression by the U [...] lso a vivid demonstration of the principle of "steadfastly managing our own affairs well." & As we all know, Huawei's chips and 5G technology have long been suppressed by sanctions fr [...] g systems.This is also a vivid manifestation of "unswervingly handling one's own affairs." & 0.9708 & 0.9772 & 0.9827 & 0.9745 \\
			\hline
			3 & 4 & It is crucial to fully recognize that only by upholding the vision of common security can  [...] n earnestly and fully implemented, both Iran and Israel would clearly be much safer today. & It is crucial to fully recognize that only by upholding the concept of common security can [...] d fully implemented, both Iran and Israel would clearly be much safer than they are today. & It is crucial to fully recognize that only by adhering to the concept of common security c [...] nd seriously implemented, Iran and Israel would obviously be much safer than they are now. & It is crucial to fully recognize that adhering to the concept of common security is essent [...] fully implemented, both Iran and Israel would undoubtedly be much safer than they are now. & It is crucial to fully realize that only by adhering to the concept of common security can [...] d fully implemented, both Iran and Israel would obviously be much safer than they are now. & 0.9729 & 0.9855 & 0.9600 & 0.9757 \\
			\hline
		\end{tabular}
	}
	\caption{Highest Semantic Similarity Verses in Global Times (Part 1)}
	\label{tab:globaltimes_highest_part1}
\end{table*}

\begin{table*}[htbp!]
	\centering
	\renewcommand{\arraystretch}{1.3}
	{\fontsize{8}{10}\selectfont
		\begin{tabular}{|p{0.5cm}|p{0.5cm}|p{2.0cm}|p{2.0cm}|p{2.0cm}|p{2.0cm}|p{2.0cm}|p{0.9cm}|p{0.9cm}|p{0.9cm}|p{0.9cm}|}
			\hline
			\textbf{Ch.} & \textbf{V.} & \textbf{Human} & \textbf{DeepSeek} & \textbf{GPT-4} & \textbf{GPT-4o} & \textbf{Google} & \textbf{H-G} & \textbf{H-DS} & \textbf{H-G4} & \textbf{H-G4o} \\
			\hline
			4 & 4 & China and Central Asian countries are companions on the path toward modernization. Since P [...] he smiles of the people, giving real, tangible warmth to the phrase "win-win cooperation." & China and Central Asian countries are fellow travelers on the path to modernization. Since [...] s on the faces of their people, giving the phrase “win-win cooperation” a tangible warmth. & China and Central Asian countries are companions on the road to modernization. Since Presi [...] ple of both sides, making the four words "cooperation and win-win" have a tangible warmth. & China and Central Asian countries are companions on the path to modernization. Since Presi [...]  on the faces of the people, giving the words "cooperation and win-win" a tangible warmth. & China and Central Asian countries are fellow travelers on the road to modernization.Since  [...] es of the people on both sides, making the four words "win-win cooperation" more tangible. & 0.9812 & 0.9916 & 0.9902 & 0.9924 \\
			\hline
			5 & 7 & Over the past five months, the Chinese and US economic and trade teams have held four roun [...] . The "big stick" in Washington's hand is nothing but a paper tiger to the Chinese people. & Over the past five months, the economic and trade teams of China and the United States hav [...] lled "big stick" wielded by Washington is nothing but a paper tiger to the Chinese people. & Over the past more than 5 months, the China-US economic and trade teams have held four rou [...] ion, the big stick in Washington's hand is nothing more than a paper tiger to the Chinese. & Over the past five months, the China-U.S. economic and trade teams have held four rounds o [...] e big stick in Washington's hand is nothing more than a paper tiger to the Chinese people. & Over the past five months, the Sino-US economic and trade teams have held four talks and r [...] re and blackmail. The big stick in Washington's hand is just a paper tiger to the Chinese. & 0.9805 & 0.9802 & 0.9862 & 0.9840 \\
			\hline
			6 & 6 & The civilizational subjectivity of China Studies lies in the fact that China's development [...]  understand China's constructive role in shaping the world from a pluralistic perspective. & The civilizational subjectivity of China studies lies in the fact that China’s development [...]  understand China’s constructive significance to the world from a pluralistic perspective. & The subjectivity of Chinese civilization lies in the fact that China's development experie [...] rstand the constructive significance of China to the world from a pluralistic perspective. & The subjectivity of Chinese civilization lies in the fact that China's development experie [...] understand the constructive significance of China to the world from a diverse perspective. & The civilized subjectivity of China studies lies in the fact that China’s development expe [...] ruly understand China's constructive significance to the world from multiple perspectives. & 0.9685 & 0.9917 & 0.9766 & 0.9825 \\
			\hline
		\end{tabular}
	}
	\caption{Highest Semantic Similarity Verses in Global Times (Part 2)}
	\label{tab:globaltimes_highest_part2}
\end{table*}

\end{CJK*} 
\end{document}